%% file: main.tex
\documentclass{article}
\usepackage{preprint,times}

\usepackage{transparent}
\usepackage{adjustbox} 
\usepackage{float}
\usepackage{calc}
\usepackage{threeparttable}
\usepackage{multirow}
\usepackage{rotating}
\usepackage{bbm}
\usepackage[utf8]{inputenc}
\usepackage[T1]{fontenc}
\usepackage{url}
\usepackage{booktabs}
\usepackage{amsfonts}
\usepackage{nicefrac}
\usepackage{microtype}
\usepackage{subcaption}
\usepackage{xr}
\usepackage{array}

\newcolumntype{L}{>{\raggedright\arraybackslash}X}
\newcolumntype{C}{>{\centering\arraybackslash}X}
\newcolumntype{R}{>{\raggedleft\arraybackslash}X}

\usepackage{setspace}
\usepackage{relsize}
\usepackage{tikz}
\usetikzlibrary{arrows,shapes,positioning,fit,decorations.pathreplacing}

\usepackage{xcolor}

\newsavebox{\myboxtable}

\usepackage{hyperref}
\hypersetup{
    colorlinks=true,
    linkcolor=BlueViolet,
    anchorcolor=black, 
    citecolor=BlueViolet
}

\input{scripts/configs}

\title{Generative Actor Critic}

\author{Aoyang Qin$^{1, 2, \star}$, Deqian Kong$^{3}$, Wei Wang$^{2}$, Ying Nian Wu$^{3}$, Song-Chun Zhu$^{1,2,4}$, Sirui Xie$^{5, \dagger, \star}$ \\
$^{1}$Department of Automation, Tsinghua University \\
$^{2}$Beijing Institute of General Artificial Intelligence (BIGAI)\\
$^{3}$Department of Statistics and Data Science, UCLA \\
$^{4}$Institute for Artificial Intelligence, Peking University \\
$^{5}$Department of Computer Science, UCLA \\
}

\begin{document}

\maketitle

\begin{abstract}
Conventional Reinforcement Learning (RL) algorithms, typically focused on estimating or maximizing expected returns, face challenges when refining offline pretrained models with online experiences. This paper introduces Generative Actor Critic (GAC), a novel framework that decouples sequential decision-making by reframing \textit{policy evaluation} as learning a generative model of the joint distribution over trajectories and returns, $p(\tau, y)$, and \textit{policy improvement} as performing versatile inference on this learned model. To operationalize GAC, we introduce a specific instantiation based on a latent variable model that features continuous latent plan vectors. We develop novel inference strategies for both \textit{exploitation}, by optimizing latent plans to maximize expected returns, and \textit{exploration}, by sampling latent plans conditioned on dynamically adjusted target returns. Experiments on Gym-MuJoCo and Maze2D benchmarks demonstrate GAC's strong offline performance and significantly enhanced offline-to-online improvement compared to state-of-the-art methods, even in absence of step-wise rewards.
\end{abstract}

\blfootnote{
        $^\star$Correspondence. $^\dagger$Now at Google DeepMind. 
        Code available at \href{https://github.com/qayqaq/Generative-Actor-Critic}{github.com/qayqaq/Generative-Actor-Critic}
}

\section{Introduction}\label{sec:intro}
\input{scripts/intro}

\section{Preliminaries}\label{sec:pre}
\input{scripts/pre}

\section{Generative Actor Critic (GAC)}\label{sec:method}
\input{scripts/method}


\section{Experiments}\label{sec:expr}

\input{scripts/exp}

\section{Discussion}\label{sec:limit}
\input{scripts/discussion}




\section*{Ethics Statement}
This work focuses on developing a fundamental algorithmic framework for reinforcement learning. The research is methodological in nature and does not involve human subjects, sensitive personal data, or direct deployment in real-world, safety-critical applications. Our contributions are confined to improving core algorithmic aspects of sequential decision-making, and the experiments are conducted in standard, simulated benchmark environments (Gym-MuJoCo and Maze2D). We do not introduce new datasets that could raise concerns regarding privacy, bias, or misuse. While we recognize that advances in reinforcement learning can have broader societal impacts when applied in downstream applications, our work does not directly engage with these deployment scenarios. The improvements described in this paper are intended for academic research and do not inherently facilitate manipulation, deception, or other unethical uses of AI agents. Overall, we believe that our research poses no direct ethical or societal risks and aligns with the principles of responsible and transparent AI development.

\section*{Reproducibility Statement}

The findings presented in this paper are supported by a detailed disclosure of our methodology and experimental setup, designed to enable full reproducibility. The core algorithmic formulation of our Generative Actor-Critic (GAC) framework is presented in §\ref{sec:method}. Our complete experimental protocol, which covers the datasets, evaluation benchmarks, and baselines, is detailed in §\ref{sec:expr}. All implementation details requisite for replication, including model architectures, training procedures, and key hyperparameters, are thoroughly documented in an appendix. Taken together, the paper and its appendix provide a complete blueprint for reproducing our results. We are committed to open science and will release the full source code upon the acceptance of the paper.

\section*{The Use of Large Language Models}
We used Gemini and ChatGPT as writing assistance tools for language polishing, grammar correction, and improving clarity. These models played no role in research conception, algorithm development, experimental design, or results generation. All scientific content, mathematical derivations, and experimental findings are original author work.

\bibliography{iclr2026_conference}
\bibliographystyle{iclr2026_conference}

\appendix
\newpage

\section{Related Work}\label{appx:related}
\input{scripts/supp_related}

\section{More Details on Experiments}\label{appx:exp}
\input{scripts/supp_exp}

\newpage
\section{Detailed Analysis of Latent Space Structure}\label{appx:latent}
\input{scripts/supp_latent}

\end{document}

%% file: scripts/configs.tex
\usepackage[utf8]{inputenc} 
\usepackage[T1]{fontenc}    
\usepackage{graphicx}
\usepackage{amsmath,amssymb,amsthm,mathabx}
\usepackage{booktabs}
\usepackage{algorithmic}
\usepackage[linesnumbered,ruled,vlined]{algorithm2e}
\usepackage{acronym}
\usepackage{enumitem}
\usepackage{nicefrac}
\usepackage{setspace}
\usepackage[skip=3pt,font=small]{subcaption}
\usepackage[skip=3pt,font=small]{caption}
\usepackage[dvipsnames,table]{xcolor}
\usepackage[capitalise]{cleveref}
\usepackage{tabularx,multirow,array,makecell}
\usepackage{overpic,wrapfig}
\usepackage{mathtools}

\usepackage{listings}
\lstdefinestyle{mystyle}{
  backgroundcolor=\color{backcolour},   commentstyle=\color{codegreen},
  keywordstyle=\color{magenta},
  numberstyle=\tiny\color{codegray},
  stringstyle=\color{codepurple},
  basicstyle=\ttfamily\footnotesize,
  commentstyle=\color{red!10!green!70}\textit,
  breakatwhitespace=false,         
  breaklines=true,                 
  captionpos=b,                    
  keepspaces=true,                 
  numbers=left,                    
  numbersep=5pt,                  
  showspaces=false,                
  showstringspaces=false,
  showtabs=false,                  
  tabsize=2
}

\definecolor{gray}{gray}{0.9}
\definecolor{tgray}{gray}{0.5}
\definecolor{mygold}{HTML}{DAA520}
\definecolor{mydeepgreen}{HTML}{006400}
\makeatletter
\DeclareRobustCommand\onedot{\futurelet\@let@token\@onedot}
\def\@onedot{\ifx\@let@token.\else.\null\fi\xspace}
\def\eg{\emph{e.g}\onedot} 

\def\ie{\emph{i.e}\onedot}

\makeatother

\makeatletter
\renewcommand{\paragraph}{%
  \@startsection{paragraph}{4}%
  {\z@}{0ex \@plus 0ex \@minus 0ex}{-1em}%
  {\hskip\parindent\normalfont\normalsize\bfseries}%
}
\makeatother

\newcommand\blfootnote[1]{%
  \begingroup
  \renewcommand\thefootnote{}\footnote{#1}%
  \addtocounter{footnote}{-1}%
  \endgroup
}
\graphicspath{{figures/}}

\crefname{algocf}{alg.}{algs.}
\Crefname{algocf}{Algrithm}{Algrithm}

\definecolor{gblue}{HTML}{4285F4}
\definecolor{gred}{HTML}{DB4437}

\acrodef{lebm}[LEBM]{Latent-space Energy-Based Model}
\acrodef{mle}[MLE]{Maximum Likelihood Estimation}
\acrodef{mcmc}[MCMC]{Markov Chain Monte Carlo}
\acrodef{mdp}[MDP]{Markov Decision Process}
\acrodef{nmdp}[nMDP]{non-Markov Decision Process}
\acrodef{pomdp}[POMDP]{Partially Observable Markov Decision Process}
\acrodef{irl}[IRL]{Inverse reinforcement learning}
\acrodef{mcts}[MCTS]{Monte Carlo Tree Search}
\acrodef{td}[TD]{Temporal Difference}
\acrodef{bco}[BCO]{Behavior Cloning from Observations}
\acrodef{ilfo}[ILfO]{Imitation Learning from Observations}
\acrodef{vi}[VI]{Variational Inference}
\acrodef{sac}[SAC]{soft Actor-Critic}
\acrodef{lanmdp}[LanMDP]{Latent-action non-Markov Decision Process}
\acrodef{lhbg}[LHBG]{Latent Hybrid BERT-GPT}

%% file: scripts/intro.tex
A central objective in sequential decision-making is to maximize \textit{the expected returns on trajectories}~\citep{sutton1998reinforcement}, denoted as $\mathbb{E}_{p(\tau)}[Y(\tau)]$, where each trajectory $\tau$ consists of a sequence of states and actions, and $Y$ is a utility function assigning return $y$ to each trajectory. Conventional Reinforcement Learning (RL) algorithms are designed to estimate and optimize this expectation during their training phase. The widely-adopted Actor-Critic framework~\citep{konda1999actor, haarnoja2018soft}, for instance, exemplifies this by learning a \textit{critic} to evaluate expected returns and an \textit{actor} to refine the policy towards maximizing these returns. This expectation-centric paradigm is particularly well-suited for online learning settings common in traditional RL~\citep{sutton1998reinforcement}, due to its amenability to efficient, iterative updates from agent-environment interactions. However, in the context of modern Generative AI (GenAI) pipelines---often characterized by an extensive offline pre-training stage on vast datasets preceding online improvement---we propose to move beyond this expectation-centric approach. We advocate for decoupling the decision modeling process into two distinct phases: (1) train-time generative modeling of the joint distribution over trajectories $\tau$ and their corresponding returns $y$, $p(\tau,y)$, and (2) test-time decision-making, framed as an inference query based on this learned joint distribution. In this paper, we introduce this approach as the Generative Actor-Critic (GAC) framework.

The Generative Actor-Critic framework fundamentally shifts the paradigm from estimating trajectories' expected returns to learning a comprehensive distribution of trajectories and their returns. This holistic approach offers several key advantages. In particular, modeling the distribution $p(\tau,y)$ naturally entails a generative \textit{critic} $p(y|\tau)$ and thus achieves \textit{generalized policy evaluation}; this perspective underscores the ability of GAC to utilize data from various sources when modeling the correlations between behaviors and outcomes, making GAC particularly well-suited for scenarios involving offline pre-training followed by online improvement. Furthermore, by explicitly modeling the entire distribution, GAC can capture complex, multi-modal relationships between behaviors and outcomes---a capacity often limited in methods focusing solely on expectation. While this advantage is shared by distributional RL~\citep{bellemare2017distributional}, the latter typically learns a return distribution conditioned on state-action pairs (often via a distributional Bellman equation) and subsequently falls back to the expectation of this learned return distribution for policy extraction~\citep{bellemare2017distributional, barth2018distributed, gruslys2018reactor}. In contrast, GAC enables a fundamental shift that replaces expectation-based \textit{actor} with a versatile inference process on $p(\tau|y)p(y)$. For instance, one could query the learned model to identify trajectories that maximize expected future returns, similar to policy optimization in distributional RL~\citep{barth2018distributed}. Alternatively, one could steer the generative process by conditioning on desired high returns to sample corresponding trajectories, akin to mechanisms in Decision Transformer (DT)~\citep{chen2021decision, zheng2022online} and Diffuser~\citep{janner2022planning, ajay2023conditional}. One could also sample diverse yet high-performing trajectories~\citep{lee2022multi} to achieve \textit{generalized policy improvement}. 

\begin{figure}[t!]
\centering
\vspace{-30pt}
\centerline{\includegraphics[width=0.4\columnwidth]{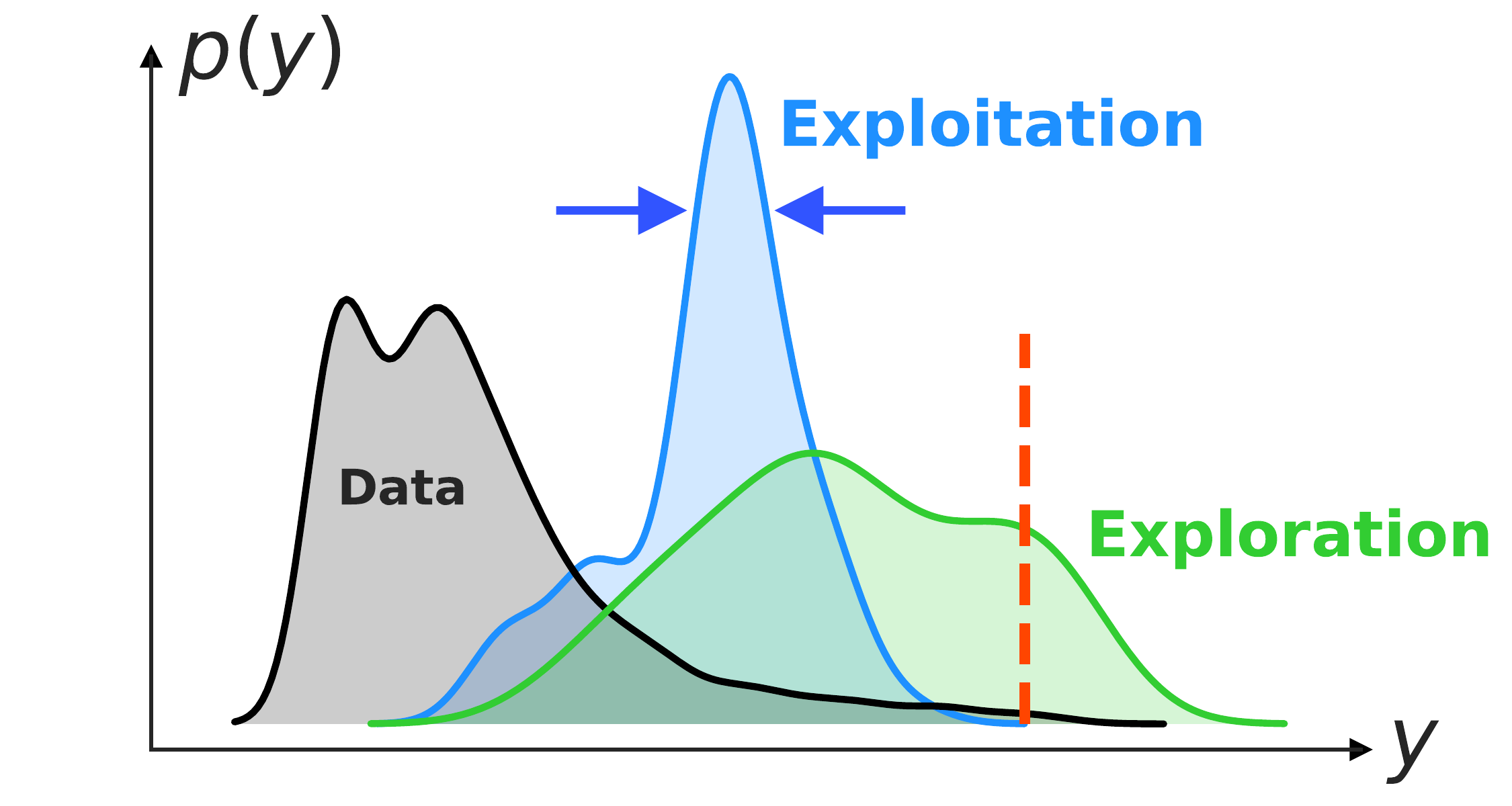}}
\caption{\textbf{Conceptual illustration of exploitation and exploration.} After modeling the data distribution (gray), GAC supports distinct test-time inference queries for various decision-making objectives. For \textit{exploitation }(blue), the objective is to maximize the expected return, leading to a focused, low-variance policy that targets high-certainty outcomes. For \textit{exploration} (green), the generative model is conditioned on a target distribution shifted towards higher returns, guiding the search for novel and potentially superior trajectories.}
\label{fig:concept}
\vspace{-15 pt}
\end{figure}

To operationalize the Generative Actor Critic framework and realize its potential, particularly for complex decision-making tasks and effective offline-to-online improvement, this paper introduces several core methodological contributions: First, we instantiate GAC using a latent-variable model structured as $p(\tau, y, z)=p(\tau|z)p(y|z)p(z)$. Inspired by recent advances like Latent Plan Transformer (LPT)~\citep{kong2024latent}, a continuous latent plan $z$ is introduced to effectively capture the correlation between high-dimensional trajectories $\tau$ and their corresponding low-dimensional returns $y$. Building on LPT's efficient inference in the latent space, we introduce non-trivial architectural and algorithmic modifications to significantly boost offline performance. Second, leveraging this latent structure, we design designated inference queries for exploitation and exploration. \textit{Exploitation} is formulated as maximizing the expected return $\mathbb{E}[y|z]$ by performing gradient ascent directly in the latent plan space. For \textit{exploration}, we propose to sample $z$ from the posterior $p(z|y^+)$ of target returns from $p(y^+)$, which is slightly shifted from $p(y)$ by a target improvement. 
\cref{fig:concept} illustrates the intuitions: Starting from the data distribution, exploitation is to refine the policy to focus on a narrow distribution of high-certainty returns, while exploration is to actively seek novel, potentially superior outcomes by shifting the target return distribution towards uncharted, higher-value regions.

Empirically, GAC achieves competitive performance on Gym-MuJoCo and Maze2D benchmarks. Our framework demonstrates strong offline learning capabilities and significantly enhanced improvement in offline-to-online scenarios compared to state-of-the-art baselines, even when step-wise rewards are absent. This strong performance is underpinned by the consistency of our generative components: the actor can be reliably steered by conditioning on target returns, while the critic provides faithful return predictions for given trajectories. Furthermore, our analysis reveals that GAC learns sophisticated internal representations in its latent space, including an implicit world model and a structured cognitive map of the environment, which underpins its strong planning capabilities.

%% file: scripts/pre.tex
\paragraph{Actor-Critic} The conventional decision-making process in RL is Markov Decision Process (MDP)~\citep{sutton1998reinforcement}, represented by a tuple $\mathcal{M} = \langle S, A, P, \pi, R, \rho, T\rangle$, comprising a state space $S$, action space $A$, transition function $P: S\times A\mapsto \Pi(S)$, policy $\pi: S\mapsto \Pi(A)$, reward function $R: S\times A\mapsto\mathbb{R}$, initial state distribution $\rho: \Pi(S)$, and horizon $T$. The decision-making objective is to maximize $\mathbb{E}[y]$, \ie, the expectation of \emph{return} $y\coloneqq\sum_{t=0}^T R(s_t, a_t)$ accumulated along the trajectories $\tau = [s_0, a_0, s_1, a_1, \ldots, a_{T-1}, s_T]$. One of the most useful identities in this formulation is the Bellman equation: if we define $Q(s_t, a_t) \coloneqq R(s_t, a_t)+\mathbb{E}_{P, \pi}[\sum_{k=1}^{T-t} R(s_{t+k}, a_{t+k})]$, we will have $Q(s_t, a_t)=R(s_t, a_t)+\mathbb{E}_{P, \pi}[Q(s_{t+1}, a_{t+1})]$. This Bellman equation, as well as its various variants~\citep{sutton1998reinforcement, ziebart2010modeling, fox2016taming, haarnoja2018soft}, provides a foundation for learning a value function $Q$ for the expectation of \emph{return-to-go} at each step $RTG_t \coloneqq \sum_{k=0}^{T-t} R(s_{t+k}, a_{t+k})$. Subsequently, the policy $\pi(a_t|s_t)$ can be updated along an approximated gradient direction $\nabla_{a_t} Q(s_t, a_t)$. That gives us the Actor-Critic framework~\citep{konda1999actor, silver2014deterministic}, where the \emph{critic} $Q$ estimates $\mathbb{E}[RTG_t|s_t, a_t]$, and the \emph{actor} $\pi$ optimizes the expectation from the critic. 

\paragraph{Distributional RL}
\citet{bellemare2017distributional} proposed to shift from the expectation-centric Bellman equation to the distributional version, $RTG_t|s_t, a_t=R(s_{t}, a_{t}) + RTG_{t+1}|s_{t+1}, a_{t+1}$. The fundamental difference is that $RTG_t$ is a random variable, while the previous object of interest, $Q_t$, is its mean. Formally speaking, in distributional RL we model $p(RTG_t|s_t, a_t)$ when training the critic, and only calculate the expectation $\mathbb{E}[RTG_t|s_t, a_t]$ when extracting policy for the actor~\citep{bellemare2017distributional, barth2018distributed, gruslys2018reactor}. \citet{bellemare2017distributional} showed that such a generative critic could express the multi-modality of $p(RTG_t|s_t, a_t)$ caused by \emph{state aliasing}~\citep{mccallum1996reinforcement}, \ie, ambiguity of hidden states under insufficient memory context. However, \citet{bellemare2017distributional} also formally derived that greedy policy selection (\ie policy optimization toward $\nabla_{a_t}\mathbb{E}[RTG_t|s_t, a_t]=0$) leads to unstable value iteration and does not guarantee contraction. 

\paragraph{RL via Sequence Modeling} Denoting the policy optimization result as $p(a_t|s_t, Q(s_t, a_t)=Q^*_{s_t})$, where $Q^*$ is the optimal value, we obtain a generative perspective for the actor. This is what underlies the idea of return-conditioned behavior cloning/sequence modeling~\citep{srivastava2019training, chen2021decision, emmons2021rvs}. Among them, Decision Transformer~\citep{chen2021decision} proposed a generalized form of policy $\pi(a_t|s_{\leq t},a_{<t}, RTG_{\leq t})$ that is learned with offline data similar to the autoregressive training of language models. However, in test time, since the model does not have access to the $RTG$s anymore, $RTG_0$ is set to be a particular high value (\eg, the maximum or a quantile above the mean) in the offline data, and is updated as $RTG_{t+1}=RTG_t-r_t$ after observing rewards. To set more plausible targets, \citet{lee2022multi} proposed to model not only the action, but also the rewards and $RTG$s. This connects the generative actor $\pi(a_t|s_{\leq t},a_{<t}, RTG_{\leq t})$ with the generative critic $p(RTG_t|s_{\leq t},a_{\leq t}, RTG_{< t})$ in distributional RL and is closely related to our GAC framework. The difference is that GAC does not assume step-wise rewards, which is argued by \citet{kong2024latent} as a more naturalistic setup that results in decision-making models that can do \emph{trajectory stitching}~\citep{fu2020d4rl}. We will pinpoint a principled synergy between the generative critic and the generative actor for online policy improvement. 

\paragraph{Offline-to-Online RL} Offline-to-online RL~\citep{wu2022supported, fujimoto2021minimalist, lyu2022mildly, kostrikov2021offline, li2023proto, beeson2022improving, zhang2023policy, nakamoto2023cal} aims to enhance the sample efficiency of online fine-tuning by leveraging offline pre-training. However, conventional expectation-centric approaches often face challenges during this transition, as their critics can become unreliable due to the action distribution mismatch between the static offline dataset and the evolving online policy~\citep{lee2022offline}. Efforts to mitigate this mismatch~\citep{kumar2020conservative, nakamoto2023cal} can, in turn, inadvertently curtail the exploratory capabilities of the learned actor. Generative actors, such as Decision Transformer, can be fine-tuned using a replay buffer that combines offline and online experiences~\citep{zheng2022online}; yet, they often lack stable and scalable mechanisms for targeted exploration in the online phase. In contrast, GAC addresses this gap by providing a distributional perspective for intentional exploration: $p(\tau|y^+)p(y^+)$. 

%% file: scripts/method.tex
GAC is a framework that separates generative decision-modeling and decision-making as inference. In \cref{sec:inference}, we introduce several probabilistic inference queries for different decision-making strategies, motivating the employment of latent-variable generative models. In \cref{sec:modeling}, we introduce a learning algorithm for GAC based on a latent-variable model~\citep{kong2024latent}. In \cref{sec:online_finetune} we describe the interplay between the actor, the critic, and the replay buffer~\citep{mnih2015human, schaul2015prioritized} in realizing exploratory online improvement. 

\subsection{Decision-Making as Inference}\label{sec:inference}

As GAC features a generative model of $p(\tau, y)$, decision-making can be framed as principled inference over it. Previous works~\citep{ziebart2010modeling, botvinick2012planning, levine2018reinforcement, abdolmaleki2018maximum, qin2023learning} in \textit{decision-making as inference} mainly take a stepwise perspective, in which the policy can be viewed as amortized variational distribution $q_{y^*}(a_t|s_t)$ for the ground-truth posterior $p(a_t|s_t, y=y^*)$, where $y^*$ denotes the optimal return. However, this posterior, along with its generalized versions in DT~\citep{chen2021decision, zheng2022online}, Diffuser~\citep{janner2022planning, ajay2023conditional}, LPT~\citep{kong2024latent}, is believed to have an \emph{optimism bias}~\citep{ziebart2010modeling, levine2018reinforcement}. Consider the decision of whether to buy a lottery ticket, where $y^*=\$10M$ but it is almost impossible to get. The expression $p(a|s,y=y^*)$ asks: what is the distribution over actions that I should take \emph{given that I have won the lottery}, to which the answer is "buy the winning ticket again", which is overly optimistic. The problem is that the event $y=y^*$ is a counterfactual, so we should not condition on it. \citet{ziebart2010modeling} attempted to bypass this issue by structuring the variational distribution of $p(\tau|y=y^*)$ as $\prod_{0}^{T-1}q_{y^*}(a_t|s_t)P(s_{t+1}|s_t, a_t)$, where $P(s_{t+1}|s_t, a_t)$ is the ground-truth world dynamics which is not conditioned on the counterfactual. 

We want to advocate for a more general solution to optimism bias. The issue is not in the inference process, but in the inference objective. As we all know, the most intuitive decision-making objective is to maximize the expected return $\mathbb{E}[y|\tau]$. And the actual optimal decision ``not to buy'' is optimal under this objective. Obviously, to a generative model $p(\tau, y)$, $\max_\tau\mathbb{E}[y|\tau]$ is a inference query distinct from $p(\tau|y=y^*)$. 

Solving $\max_\tau\mathbb{E}[y|\tau]$ directly would employ gradient ascent that requires costly backpropagation through time. This is where latent modeling excels. Consider a generative model with continuous latent variables $z$
\begin{equation}\label{eq:joint_p}
    \begin{aligned}
        p_\theta(\tau,y,z) & = p_\alpha(z)p_\beta(\tau|z)p_\gamma(y|z), \\
    \end{aligned}
\end{equation}
where conditional independence is assumed between the trajectory $\tau$ and its return $y$, positioning $z$ as an information bottleneck. Intuitively, the latent $z$ are \emph{plans} that abstract trajectories around their returns. As long as $p_\gamma(y|z)$ and $\mathbb{E}[y|z]$ have analytical forms, gradient-based inference can be lifted to the variational posterior $q_\phi(z)$ in the continuous latent space. To ground the inferred plans within the learned behavior manifold, we regularize the exploitation objective with a KL to the prior $p_\alpha(z)$:
\begin{equation}\label{eq:infer_exploit}
    \max_{\phi} \mathbb{E}_{q_\phi(z)}\left[\mathbb{E}[y|z]\right] - D_\text{KL}(q_\phi(z)||p_\alpha(z)).
\end{equation}
Once $z$ is inferred, the policy is entailed by $p_\beta(\tau|z)$. In fact, $z$ can be inferred at any time steps, which is particularly useful when the environment dynamics are stochastic and the policies drift from plans. If we only infer the $z$ at the initial state, we obtain an open-loop policy; if we keep updating $z$ to incorporate the generated partial sequence $\tau_t=[s_0, a_0, s_1, a_1, \ldots, a_{t-1}, s_t]$ via:
\begin{equation}\label{eq:infer_replan}
    \max_{\phi} \mathbb{E}_{q_\phi(z)}\left[\mathbb{E}[y|z]\right]-D_\text{KL}(q_\phi(z)||p_\alpha(z|\tau_t)),
\end{equation}
where $p_\theta(z|\tau_t) \propto p_\alpha(z) p_\gamma(\tau_t|z)$, we obtain a policy from closed-loop replanning.

Continuous latent modeling also enables other inference queries that would not be imaginable otherwise. In fact, the optimization in \cref{eq:infer_exploit} can be viewed as a special type of joint sampling over $p(z, y)$. Returning to the lottery example, a decision policy $p(z|y)$ would not be unreasonable if we had more nuanced control over optimism. In other words, if, instead of sampling from a fixed target $y^*$, we sample from a marginal distribution $p(y^+)$ that is slightly shifted right from $p(y)$, the \emph{optimism} in $p_\theta(z|y^+)$ is tamed with the awareness of $p(y^+)$. To sample from $p_\theta(z|y^+)$, we can employ classical Variational Bayes to $\min_\phi D_\text{KL}(q_\phi(z)||p_\theta(z|y^+))$, where $p_\theta(z|y) \propto p(z) p_\gamma(y|z)$, which is equivalent to maximizing the evidence lower bound (ELBO) \citep{murphy2018machine}:
\begin{equation}\label{eq:infer_explore}
    \max_{\phi} \mathbb{E}_{q_\phi(z)}[\log p_\gamma(y^+|z)] - D_\text{KL}(q_\phi(z)||p_\alpha(z)).
\end{equation}
Hopefully, the optimized $q_\phi(z)$ is a faithful approximation of $p_\theta(z|y^+)$, from which we can sample $z$. We will introduce how to sample from $p(y^+)$ in \cref{sec:online_finetune}. 

\subsection{Generative Modeling}\label{sec:modeling}

LPT~\citep{kong2024latent} is an instantion of the latent-variable generative model desribed in \cref{eq:joint_p}. Building on their conceptualization, we made two modifications that align well with some common sense about decision-making: (1) the initial state $s_0$ is not generated from a plan, so we move it into the conditioning; (2) the latent abstractions of the trajectories and their returns are better to be distinct and associated instead of identical, so we reserve one vector $z_y$ for returns, leave the rest $z_{\setminus y}$ for trajectories, and associate them in the prior $p_\alpha(z)$. This gives us the factorization: 
\begin{equation}\label{eq:joint_p_factor}
    \begin{aligned}
        p_\theta(\tau,y,z) & = \rho(s_0) p_\alpha(z|s_0)p_\beta(\tau|s_0, z_{\setminus y})p_\gamma(y|z_y). \\
    \end{aligned}
\end{equation}
Following LPT, the prior $p_\alpha(z)$ is implemented as an implicit generative model, but we replace the UNet transformation~\citep{ronneberger2015u} in LPT with a Transformer encoder with bidirectional mask. To sample $z\sim p_\alpha(z)$, we first sample a set of isotropic Gaussian $\epsilon \sim \mathcal{N}(0, I)$ and transform them with the Transformer $z=f_\alpha(s_0, \epsilon)$. 

We also inherit LPT's trajectory generator as a $z$-conditioned autoregressive model with a finite context window of size $M$: $p_\beta(\tau|s_0, z_{\setminus y})=\prod\nolimits^{T-1}_{t=0}p_\beta(a_t|s_{\leq t},a_{<t},z_{\setminus y})p_\beta(s_{t+1}|s_{t-M: t},a_{t-M: t},z_{\setminus y})$. This design forces the latent $z$ to serve as global carriers of information, bridging temporal segments that would otherwise be disconnected due to the limited context. The generator is parameterized by a causal Transformer decoder that incorporates the latent $z$ via cross-attention. The stepwise policy and transition distributions are modeled as Gaussians with learnable or fixed variances.

The return predictor models $p_\gamma(y | z_y)$ as a Gaussian $\mathcal{N}(\mu_\gamma(z_y), \sigma_y^2)$, where $\mu_\gamma$ is an MLP that predicts $y$ from the special latent vector $z_y$. The variance $\sigma_y^2$ is either treated as a hyperparameter or learned jointly. Note that $p_\gamma(y | z_y)$ being a Gaussian would not constrain the capability of modeling multi-modal return distribution because $p(y|\tau)=\int p(y|z_y)p(z_y|\tau)\text{d}z_y$ can be very expressive. 

For a data point $(\tau, y)$, Maximum Likelihood Estimate (MLE) of latent-variable models is intractable. Instead, we introduce a variational distribution $q_\phi(\epsilon|\tau, y) = \mathcal{N}(\mu, \sigma^2)$, parameterized by $\phi=(\mu, \sigma)$, \ie, the mean vector $\mu$ and a diagonal covariance matrix $\sigma^2$~\citep{blei2017variational, jordan1999introduction}, and maximize the ELBO:
\begin{equation}\label{eq:elbo}
    ELBO(\theta,\phi)=\mathbb{E}_{q_{\phi}(\epsilon)}[\log p_\theta(\tau, y|s_0, \epsilon)]-D_\text{KL}(q_{\phi}(\epsilon)||p(\epsilon)).
\end{equation}
We use the re-parametrization trick~\citep{kingma2013auto} in $\mathbb{E}_q$. The training employs alternating update of the local parameters $\phi$ specific to each $(\tau, y)$ with classical Variational Bayes~\citep{hoffman2013stochastic} and the global parameters $\theta = (\alpha, \beta, \gamma)$ shared across all training data. \citet{kong2025latent} recently showed that this training scheme is effective at GPT-2 scale language models. 

Inference queries in \cref{eq:infer_exploit} and \cref{eq:infer_explore} can be adjusted accordingly to incorporate conditioning on $s_0$ and reparametrization with $\epsilon$. Between them, \cref{eq:infer_exploit} enables exploitation during test time, while \cref{eq:infer_explore} facilitates exploratory online fine-tuning. 

\subsection{Exploratory Online Fine-tuning}\label{sec:online_finetune}

A key advantage of the GAC framework is that the training objective—maximizing the ELBO in \cref{eq:elbo}—remains consistent for both offline pre-training and online fine-tuning. The online fine-tuning is fueled by higher-quality trajectories collected with a principled exploration strategy.

Our exploration strategy is designed to sample trajectories from an improved distribution by targeting returns slightly higher than what the model has previously achieved. Ideally, one might tilt the model's prior $p(z)$ to generate a desired $p(y^+)$. However, we adopt a simpler and more direct approach. We leverage a prioritized replay buffer, initialized with the offline dataset, which serves as an empirical return distribution $p_{\text{data}}(y)$. To generate a target $y^+$ from the target distribution $p(y^+)$, we first sample a high-performing return $y$ from the top-k quantile of the replay buffer. We then add a small, positive increment $\Delta y$ to it, creating an optimistic target $y^{+} = y + \Delta y$. This target is used to condition the inference process as described in \cref{eq:infer_explore}, yielding latent plans $z \sim p(z|y^+)$ that guide the agent to explore potentially out-of-distribution (OOD) yet superior trajectories.

While the returns collected by the actor $p(z|y)p_{\text{data}}(y)$ from the environment are generally consistent with the target $y$ sampled from the empirical distribution, the introduction of $\Delta y$ creates a mismatch between the target distribution $p(y^+)$ and the ground-truth return distribution. How to reliably set $\Delta y$ to balance exploration with stability is an interesting research question that we leave for future work. In this work, we treat $\Delta y$ as a manually tuned hyperparameter. Empirically, we find that this optimistic distributional targeting strategy effectively guides exploration; while it does not always lead to trajectories that meet the optimistic $y^+$, it consistently outperforms fixed target $y^*$ and gradually shifts the distribution of collected returns toward higher values, as shown in \cref{fig:online_results}.

The online finetuning process is structured in stages. In each stage, we first collect a set of new trajectories using the exploration policy described above. These newly collected trajectory-return pairs are then added to the replay buffer, enriching the dataset with recent, high-quality experiences. Following data collection, we fine-tune the GAC model on the updated replay buffer for a set number of steps, using the same ELBO maximization objective as in the pre-training phase. This cycle of exploration, replay buffer update, and model fine-tuning is repeated, allowing GAC to progressively adapt and improve its performance through online interaction. The offline pre-training and online fine-tuning algorithms are described by \cref{alg:offline} and \cref{alg:online} in the appendix.

%% file: scripts/exp.tex
\begin{table}[t]
\caption{\textbf{Results on various inference objectives after offline pre-training .} Comparison between the average normalized returns of GAC against several baselines with only final return signals, evaluated without any online finetuning. The mean and standard deviation of different GAC’s inference strategies are based on 100 evaluation trajectories. The results illustrate the distinct behaviors among the exploitation query (GAC-$\mathbb{E}[y]$), the exploration query (GAC-$p(y^+)$), fixed target steering (GAC-$y^*$), and sampling from the prior ($p(y|z)p(z)$). The best result in each row is highlighted in bold. In the 1st column, `-m' and `-r' are abbr. for -medium and -replay.}
\label{tab:combined_offline_results}
\begin{center}
\begin{footnotesize}
\setlength{\tabcolsep}{4.5pt}
\begin{tabularx}{\textwidth}{l|ccccc|cccc}
\toprule
\multirow{2}{*}{Dataset} & \multicolumn{5}{c|}{Baselines} & \multicolumn{4}{c}{GAC Inference Strategies} \\
\cmidrule(lr){2-6} \cmidrule(lr){7-10}
& IQL & CQL & DT & QDT & LPT & GAC-$\mathbb{E}[y]$ & GAC-$p(y^+)$ & GAC-$y^*$ & $p(y|z)p(z)$ \\
\midrule
hopper-m & 35.1 & 23.3 & 57.3 & 50.7 & 58.5 & 60.2${\pm}$8.0 & \textbf{73.2}${\pm}$13.6 & 57.1${\pm}$10.2 & 53.7${\pm}$17.0 \\
hopper-m-r & 13.9 & 7.7 & 50.8 & 38.7 & 71.2 & \textbf{80.1}${\pm}$9.2 & 67.5${\pm}$25.1 & 61.3${\pm}$23.7 & 34.3${\pm}$23.1 \\
walker2d-m & 49.1 & 0.0 & 69.9 & 63.7 & 77.8 & \textbf{80.2}${\pm}$4.6 & 78.9${\pm}$8.9 & 78.5${\pm}$6.7 & 75.6${\pm}$10.1 \\
walker2d-m-r & 5.3 & 3.2 & 51.6 & 29.6 & 72.3 & \textbf{78.9}${\pm}$9.3 & 73.3${\pm}$10.2 & 76.6${\pm}$16.2 & 34.3${\pm}$27.7 \\
halfcheetah-m & 8.5 & 1.0 & 42.4 & 42.4 & 43.1 & \textbf{43.6}${\pm}$2.5 & 41.5${\pm}$5.1 & 42.5${\pm}$1.5 & 41.2${\pm}$5.6 \\
halfcheetah-m-r & 5.2 & 7.8 & 33.3 & 32.8 & 39.6 & \textbf{39.8}${\pm}$7.8 & 38.8${\pm}$8.9 & 36.7${\pm}$8.2 & 33.0${\pm}$12.0 \\
\midrule
maze2d-umaze & 4.5 & 3.9 & 28.4 & 2.57 & 65.4 & \textbf{67.8}${\pm}$21.4 & 64.2${\pm}$22.7 & 59.2${\pm}$21.6 & 35.7${\pm}$23.4 \\
maze2d-medium & 3.5 & -3.6 & -2.4 & -2.58 & 20.6 & \textbf{74.5}${\pm}$71.0 & 63.3${\pm}$71.4 & 61.2${\pm}$70.2 & 19.9${\pm}$56.6 \\
maze2d-large & 2.0 & -1.2 & -2.5 & -2.51 & 37.2 & \textbf{50.3}${\pm}$40.4 & 39.5${\pm}$42.3 & 28.9${\pm}$39.0 & -0.4${\pm}$6.1 \\
\bottomrule
\end{tabularx}
\end{footnotesize}
\end{center}
\vspace{-15pt} 
\end{table}

We evaluate GAC on standard offline and offline-to-online reinforcement learning benchmarks from the D4RL Gym-MuJoCo suite (Halfcheetah, Hopper, and Walker2D) and Maze2D navigation task (Umaze, Medium, Large). The former tasks have dense step-wise reward, while the latter ones feature in binary step-wise reward where the agent is rewarded 1 point when it is around the goal. Our experiments are conducted in a setting where only total trajectory returns are available. The training data contains plenty of suboptimal trajectories. We compare GAC against state-of-the-art methods and analyze the effectiveness of its different inference strategies for exploitation and exploration.

Our baselines come across several paradigms: offline and offline-to-online RL (IQL~\citep{kostrikov2021offline}, CQL~\citep{kumar2020conservative}, Cal-QL~\citep{nakamoto2023cal}), sequence modeling approaches (DT~\citep{chen2021decision}, ODT~\citep{zheng2022online}, QDT~\citep{yamagata2023q}, LPT~\citep{kong2024latent}), and classic online RL (PPO~\citep{schulman2017proximal}, SAC~\citep{haarnoja2018soft}) for reference. Notably, except for LPT, these methods were designed for step-wise rewards. We adapt them to our trajectory-return-only setting for a fair comparison, highlighting the distinct advantage of models that do not rely on dense step-wise reward signals. For online fine-tuning, we use a staged training pipeline, with 100 trajectories for MuJoCo and 500 trajectories for Maze2D each stage.

\textbf{Offline Pre-training.} We report the offline pre-training results in the left panel of Table \ref{tab:combined_offline_results}, evaluated using the exploitative inference from \cref{eq:infer_exploit} (termed GAC-$\mathbb{E}[y]$). Our model consistently outperforms all baselines across the tested environments. GAC's advantage stems from its unique generative approach. Unlike conventional methods rooted in temporal difference learning, GAC forgoes explicit per-step credit assignment. Instead, it learns an implicit understanding of action-outcome relationships via cross-attention across the latent token sequence. This allows the model to learn a smooth and structured latent space of behaviors, enabling it to compose novel, high-quality trajectories by interpolating between successful sub-sequences found in the training data. The robustness of this approach is particularly evident in Maze2D benchmarks, which are characterized by sparse rewards and long-horizon planning challenges. While TD-based methods like IQL and CQL are heavily dependent on dense, step-wise rewards, GAC maintains high performance even with only trajectory-level returns. This highlights a key strength of our framework: GAC achieves superior performance despite operating without the granular, step-wise reward signals that most of these powerful baselines rely on for effective training.


\begin{figure}[t]
\vspace{-10pt}
\begin{center}
\centerline{\includegraphics[width=1.\columnwidth]{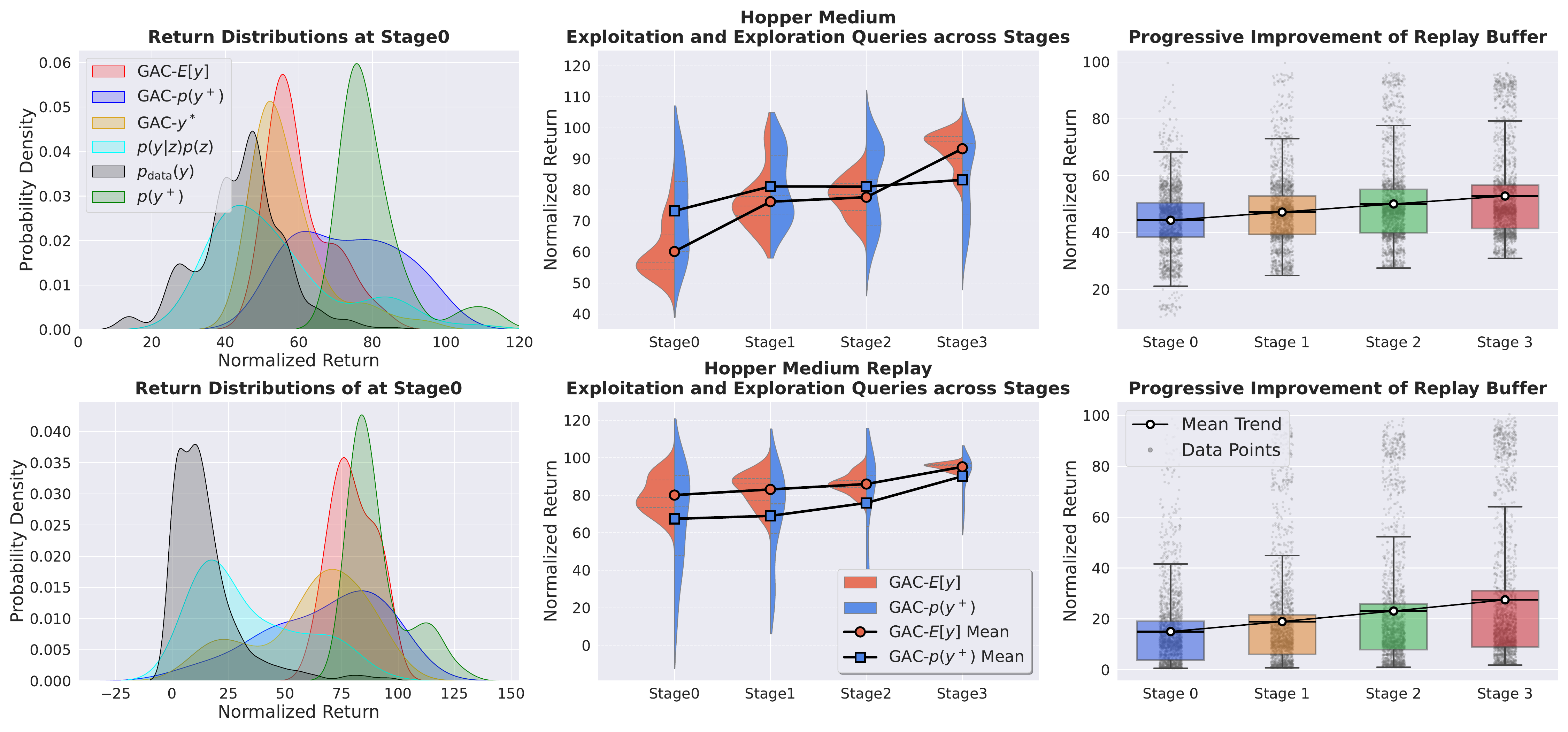}}
\caption{\textbf{Illustration of return distributions in Mujoco.} The left column displays the return distributions achieved by various inference objectives after offline pre-training, highlighting the explorative and exploitative behavior. The middle column presents a split violinplot for the inference objectives of GAC-$\mathbb{E}[y]$ and GAC-$p(y^+)$, showing the progressive performance gains and the explorative and exploitative behaviors at three sequential fine-tuning stages. The right column visualizes the progressive improvement in the datasets' quality, as evidenced by the upward shift of the dataset's return distribution over the fine-tuning stages.}
\label{fig:online_results}
\end{center}
\vspace{-30pt}
\end{figure}

\textbf{Inference Strategies.} We analyze the practical implications of the different inference strategies introduced in \cref{sec:inference}. We compare three approaches: GAC-$\mathbb{E}[y]$ for pure exploitation via latent space optimization; GAC-$y^*$, which conditions on a standard high-value target return similar to DT~\citep{chen2021decision}; and GAC-$p(y^+)$, our proposed exploration strategy that conditions on dynamically adjusted targets sampled from the replay buffer. 

The return distributions collected by actors from different inference objectives, visualized in the left column of \cref{fig:online_results} (with mean and standard deviation summarized in the right panel of \cref{tab:combined_offline_results}), highlight their distinct behaviors. The return distribution {\transparent{0.8}\textcolor{cyan}{$p(y|z)$}}, generated from random prior latent plans from $p(z)$, is dispersed by principle and effectively covers the returns seen in the training data, confirming that the model has not collapsed to a single mode. In sharp contrast, the {\transparent{0.8}\textcolor{red}{GAC-$\mathbb{E}[y]$}} strategy produces a focused, low-variance distribution concentrated on high-certainty, high-return outcomes, underscoring its efficacy as a pure exploitation method. Such exploitation effect is apparent when comparing with chasing the empirical maximum $y^*$ with {\transparent{0.8}\textcolor{mygold}{GAC-$y^*$}. To encourage exploration, we sample target returns from {\transparent{0.8}\textcolor{mydeepgreen}{$p(y^+)$}}, which apparently have better coverage at high-return and even OOD regions than GAC-$\mathbb{E}[y]$. Based on these targets, {\transparent{0.8}\textcolor{blue}{GAC-$p(y^+)$}} achieves a high mean return with greater variance than GAC-$\mathbb{E}[y]$. While GAC-$p(y^+)$'s optimistic conditioning on $y^+$ leads to a delibrate mismatch between its resulting distribution and $p(y^+)$, it reflects a realistic improvement over the $p_\text{data}(y)$, showcasing successful guided exploration toward better outcomes. As GAC-$p(y^+)$'s target distribution is anchored at the data distribution, it also outperforms GAC-$y^*$, whose reliance on a single fixed target often leads to overestimation (\textit{optimism bias}) of the outcomes. 

\textbf{Online Fine-tuning.} \cref{tab:online_results} reports the final returns compared to other methods after fine-tuning. While the gains are relatively small in some tasks, this is often a consequence of its higher initial offline performance. Notably, GAC remains highly effective even without dense, step-wise rewards—a setting where pure online RL methods like PPO and SAC typically underperform. Unlike many offline-to-online methods that rely on explicit optimism or conservatism, GAC improves by leveraging its principled exploration strategy. This strategy systematically expands toward trajectories with higher returns by conditioning on targets set slightly beyond the empirical best of the current dataset. When these exploratory rollouts yield higher-performing trajectories, they are added to the replay buffer. To visualize this dynamic process, \cref{fig:online_results} plots the return distributions for MuJoCo from different inference objectives across several fine-tuning stages. The figure clearly illustrates the distinction between exploration and exploitation, tracking the progressive performance improvement via the continuous, positive shift in the collected data's return distribution. As for Maze2D, we illustrate the online improvement through heatmaps in the top row of \cref{fig:online_results_maze} which display the mean return achieved from different starting cells across three stages of fine-tuning. We can see a clear progression that from Stage 1 to Stage 3, the number of high-return cells increases, indicating that the agent learns to solve the task from a wider array of initial states. For further details, please refer to \cref{appx:exp}.

\begin{table}[t]
\vspace{-20pt}
\caption{\textbf{Online fine-tuning results.} Comparison of normalized returns before and after online fine-tuning with only access to final return. We fine-tune GAC for 3 stages with data collected by GAC-$p(y^+)$ and report the final GAC-$\mathbb{E}[y]$ performance from 100 trajectories. The best final result for each dataset is highlighted in bold. $\delta$ denotes the performance gain over the offline pre-trained models reported in \cref{tab:combined_offline_results}.}
\label{tab:online_results}
\centering
\smaller
\setlength{\tabcolsep}{4pt}
\begin{tabularx}{\textwidth}{l|c|c|Cc|cc|Cc|Cc|Cc|cc}
\toprule
Dataset
& \multicolumn{2}{c|}{Online RL}
& \multicolumn{2}{c|}{ODT}
& \multicolumn{2}{c|}{CQL}
& \multicolumn{2}{c|}{Cal-QL}
& \multicolumn{2}{c|}{IQL}
& \multicolumn{2}{c|}{LPT}
& \multicolumn{2}{c}{GAC} \\
& PPO & SAC & online & $\delta$ & online & $\delta$ & online & $\delta$ & online & $\delta$ & online & $\delta$ & online & $\delta$ \\
\midrule
hopper-m & \multirow{2}{*}{13.1${\pm}$2.1} & \multirow{2}{*}{11.2${\pm}$1.5} & 57.6 & +0.3 & 29.6 & +6.3 & 32.8 & +9.5 & 25.0 & -10.1 & 64.8 & +6.3 & \textbf{93.3}${\pm}$5.1 & +33.1 \\
hopper-m-r &  &  & 65.2 & +14.4 & 8.4 & +0.7 & 24.1 & +16.4 & 12.6 & -1.3 & 72.4 & +1.2 & \textbf{95.2}${\pm}$1.9 & +15.1 \\
walker2d-m & \multirow{2}{*}{9.5${\pm}$1.6} & \multirow{2}{*}{4.0${\pm}$0.9} & 70.7 & +0.8 & 1.9 & +1.9 & 1.2 & +1.2 & 50.1 & +1.0 & 79.5 & +1.7 & \textbf{85.1}${\pm}$4.9 & +4.9 \\
walker2d-m-r &  &  & 57.3 & +5.7 & 0.5 & -2.7 & 3.5 & +0.3 & 6.9 & +1.6 & 79.0 & +6.7 & \textbf{85.4}${\pm}$5.3 & +6.5 \\
halfcheetah-m & \multirow{2}{*}{1.3${\pm}$0.1} & \multirow{2}{*}{1.7${\pm}$0.2} & 40.7 & -1.7 & 2.8 & +1.8 & 3.1 & +2.1 & 8.9 & +0.4 & 43.2 & +0.1 & \textbf{44.3}${\pm}$1.1 & +0.8 \\
halfcheetah-m-r &  &  & 24.4 & -8.4 & 6.4 & -1.4 & 2.3 & -5.5 & 7.4 & +2.2 & 40.6 & +1.0 & \textbf{40.9}${\pm}$0.8 & +1.1 \\
\midrule
maze2d-umaze & 55.5${\pm}$1.1 & 61.8${\pm}$1.2 & 11.2 & -17.2 & 5.8 & +1.9 & 4.9 & +1.0 & 29.3 & +24.8 & 67.2 & +1.8 & \textbf{83.5}${\pm}$23.9 & +15.7 \\
maze2d-medium & 29.3${\pm}$3.9 & 46.8${\pm}$4.7 & 2.5 & +4.9 & -1.8 & +1.8 & 0.9 & +4.5 & 23.1 & +19.6 & 26.1 & +5.5 & \textbf{166.1}${\pm}$33.4 & +91.6 \\
maze2d-large & -0.8${\pm}$0.8 & 17.7${\pm}$0.8 & 2.2 & +4.7 & 0.3 & +1.5 & 1.1 & +2.3 & 7.4  & +5.4 & 40.1 & +2.9 & \textbf{94.2}${\pm}$42.7 & +43.9  \\
\bottomrule
\end{tabularx}
\vspace{-5pt}
\end{table}

\begin{figure}[t]
\begin{center}
\centerline{\includegraphics[width=1.\columnwidth]{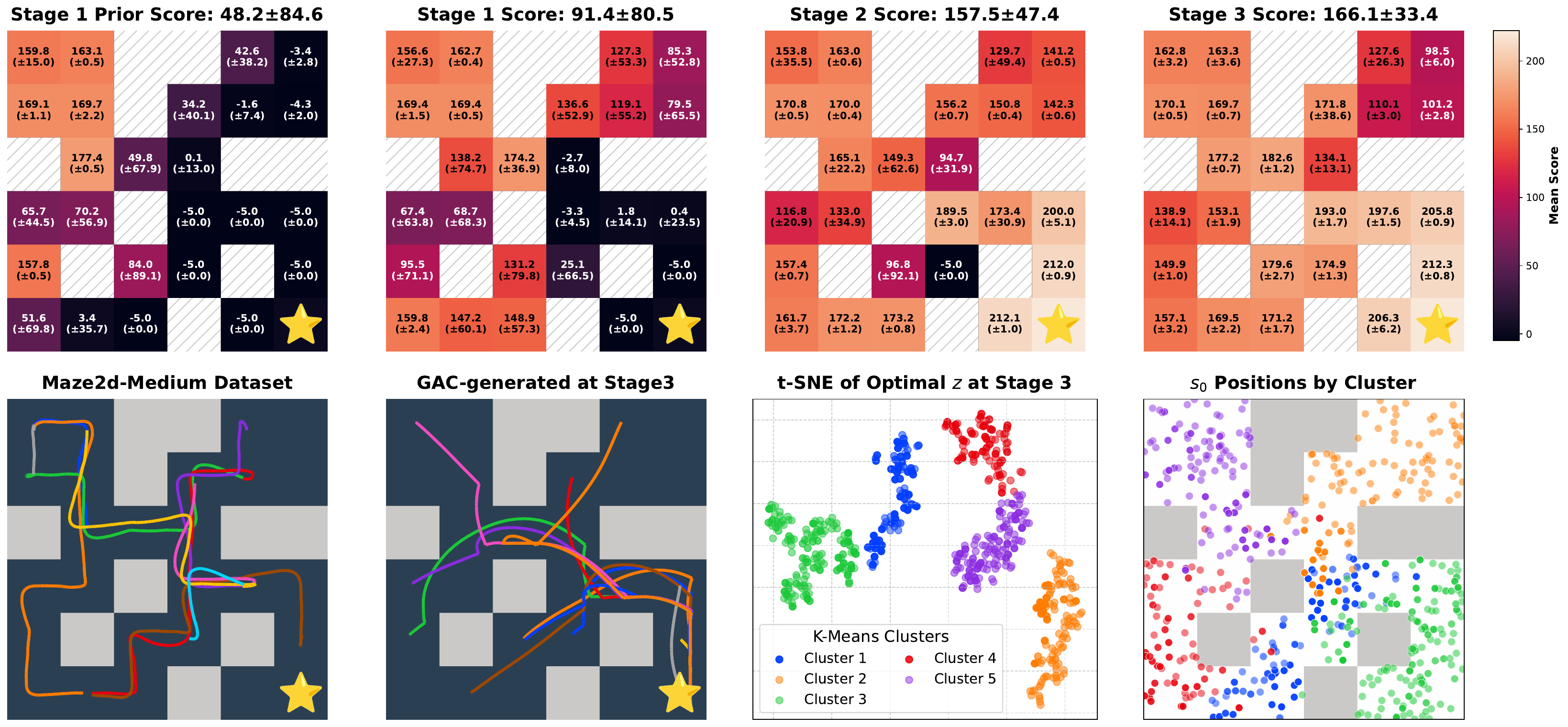}}
\caption{\textbf{Illustration of online improvement and emergent properties in Maze.} The top row heatmaps visualize performance from different starting cells. The first two panels compare inference strategies at Stage 1, demonstrating the superiority of the exploitative strategy GAC-$\mathbb{E}[y]$ (Stage 1 Score) over sampling from the prior (Stage 1 Prior Score). The progression from Stage 1 to Stage 3 (second to fourth panels) then illustrates a monotonic improvement in online fine-tuning. Subplots in the bottom row visualize the model's emergent capabilities. The model is trained on suboptimal, axis-aligned trajectories (far left). After fine-tuning, GAC generates novel, efficient diagonal shortcuts (second from left). This advanced planning is likely supported by an emergent cognitive map: a t-SNE visualization shows the latent space of plans, $z$, self-organizing into distinct clusters (third from left), which directly correspond to spatially coherent regions of starting states, $s_0$ (far right).}
\label{fig:online_results_maze}
\end{center}
\vspace{-30pt}
\end{figure}

\textbf{Emergent Properties.} More significantly, there emerge an abundant set of sophisticated properties, with two identified and illustrated in the bottom row of \cref{fig:online_results_maze}. First, GAC learns to generate novel, efficient behaviors unseen in the offline data. Despite being trained on suboptimal, axis-aligned trajectories, the fine-tuned model discovers and executes diagonal shortcuts and stops precisely at the goal. This indicates that it has internalized an implicit world model of the environment and reward structure without explicit supervision~\citep{gurnee2023language, vafa2024evaluating}. Second, GAC's latent space self-organizes into a structured representation of the environment. When we perform posterior inference to map latent plans $z$ from a dense grid of initial states $s_0$, they form distinct clusters, with each cluster corresponding to a specific spatial region. This emergent organization is strikingly analogous to hippocampal \textit{place cells}~\citep{o1976place, zhao2025place}, which implies a cognitive map~\citep{whittington2022build, tolman1948cognitive} is formed, without any spatial priors, to provide a structured foundation for GAC's advanced, long-horizon planning capabilities.

\textbf{Actor and Critic.} The success of online improvement is rooted in the efficacy of the actor and the critic learned through generative modeling. To evaluate the consistency of the actor, in \cref{fig:corr} we plot for Hopper-Medium the actual returns achieved by the actor steered by some target returns, which show a strong positive correlation. Depending on the data distributions, actors at different stages behave under different spectrums of variance. Crucially, at Stage 0, the actor can be steered to achieve OOD target returns. Even after Stage 3 when the actor's performance is almost optimal, and exceedingly high target returns won't cause a degradation (see \cref{app:ood_inference} for detailed analysis on robustness given OOD targets). There is also a strong correlation between a trajectory's predicted return and the actual return, which justifies the consistency of the critic. 

\textbf{Closed-loop replanning.} To showcase GAC's replanning capability, we designed a proof-of-concept experiment in Maze2D-Large (\cref{fig:replan}). The task requires the agent to navigate a long-range path from a starting state $A$ to a distal goal, passing through an intermediate waypoint $B$. We first observed that an open-loop policy, created by committing to a latent plan $z$ inferred at state $B$, can reliably guide the agent to the goal. However, when starting from $A$, a plan inferred at the start often fails; the agent successfully follows the optimal path to $B$ but then deviates and fails to reach the goal. Our experiment demonstrates that by manually triggering replanning—updating the latent plan $z$ by \cref{eq:infer_replan} upon reaching the vicinity of $B$—the agent can correct its trajectory. This simple act of replanning significantly increases the success rate, highlighting GAC's ability to leverage its generative nature for effective, state-aware self-correction.


\begin{figure}[t]
\vspace{-30pt}
\begin{center}
\centerline{\includegraphics[width=1.\columnwidth]{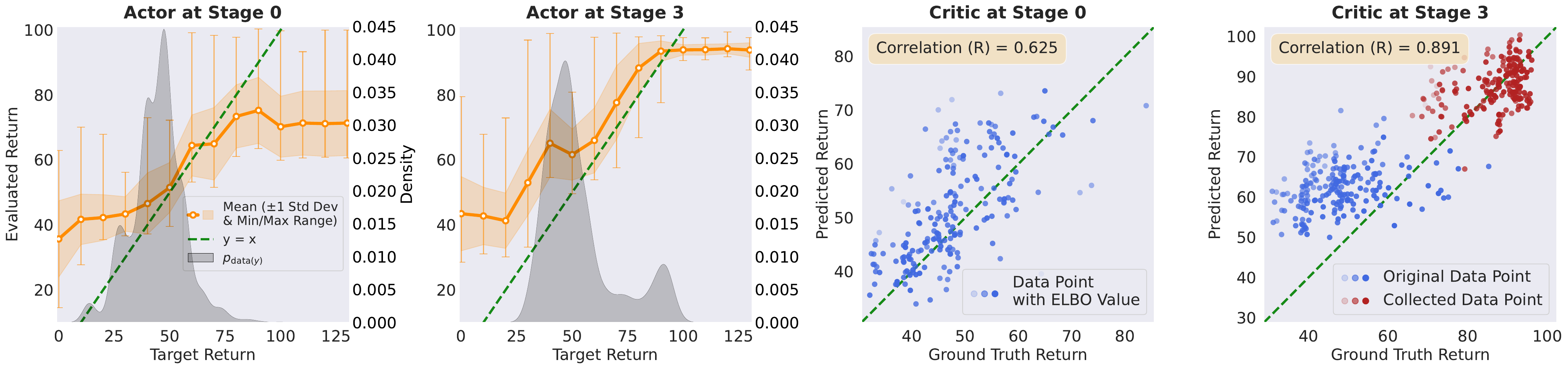}}
\caption{\textbf{Analysis of actor and critic.} The left two panels illustrate the actor's consistency. For each target return (x-axis), we infer 50 latent plans to generate corresponding trajectories, and plot the mean of their evaluated returns (y-axis). The right two panels demonstrate the critic's consistency. For each trajectory from the dataset with a ground truth return (x-axis), we infer 50 latent plans and use their average predicted return as the y-axis value. Points with higher ELBO values are less transparent and closer to the ideal $y=x$ line, indicating more reliable predictions. The Stage 3 plot distinguishes original (blue) from newly collected online data (red).}
\label{fig:corr}
\end{center}
\vspace{-10pt}
\end{figure}

\begin{figure}[t]
\vspace{-10pt}
    \begin{minipage}[b]{0.63\textwidth} 
        \makebox[0pt][l]{%
            \raisebox{-1.5cm}{
                \includegraphics[height=3.cm]{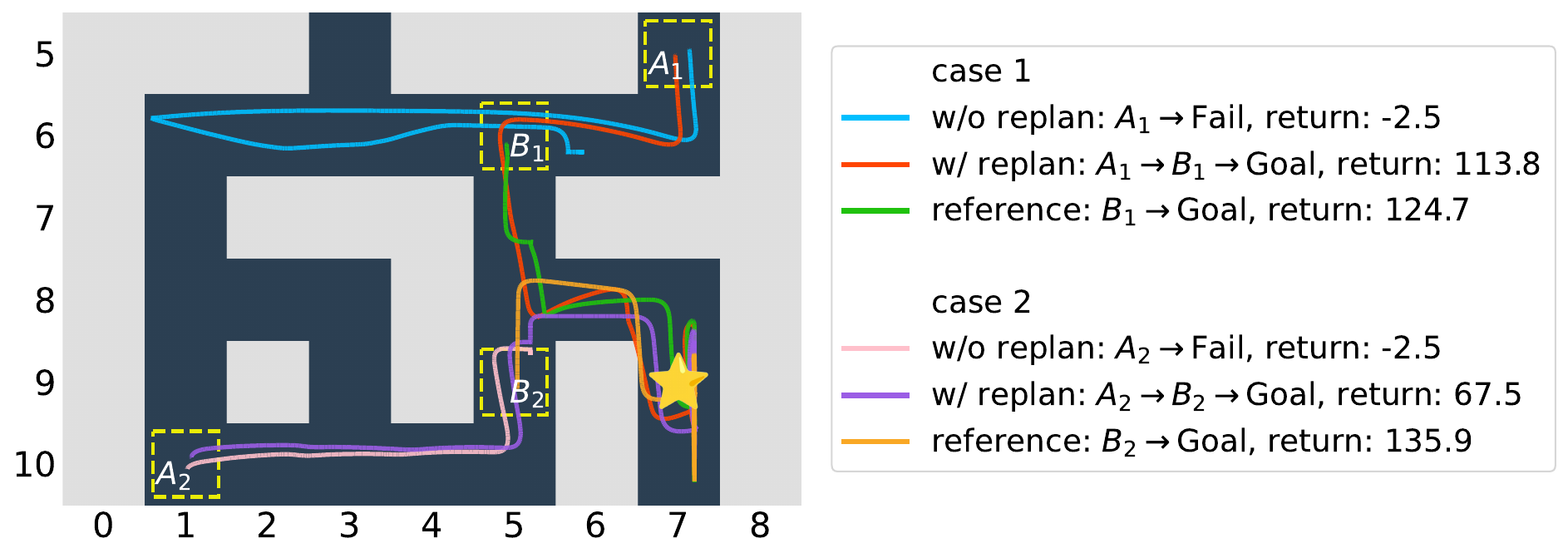}
            }%
        }
    \end{minipage}%
    \begin{minipage}[b]{4.5cm} 
        \begin{center}
        \begin{footnotesize}
        \setlength{\tabcolsep}{4pt} 
        \begin{tabularx}{\textwidth}{c|cc}
        \toprule
        case & 1 & 2 \\
        \midrule
        $A$ & (7,5) & (1,10) \\
        $B$ & (5,6) & (5,9) \\
        w/o replan & 15.4\% & 27.0\% \\
        w/ replan & 37.0\% & 46.0\% \\
        gain & +140.3\% & +70.4\% \\
        \bottomrule
        \end{tabularx}
        \end{footnotesize}
        \end{center}
    \end{minipage}%
    \captionof{figure}{\textbf{Experiment on replanning.} While the initial plan from $A$ rarely succeeds, with a replanning at the critical intermediate state $B$, the agent can correct its course towards the goal. In the left we visualize a failure and a correction starting from $A$ and a reference from $B$. In the right we report the success rate.}
 \label{fig:replan} 
 \vspace{-20pt}
\end{figure}

\textbf{Limitations.} 
While GAC demonstrates strong general performance, we note that methods with access to step-wise rewards can achieve slightly higher scores in certain tasks (\cref{tab:supp_online_results}), such as PPO in Maze2D and Cal-QL in MuJoCo. However, we argue that GAC's performance could become significantly more competitive with the development of a currently unrealized feature: autonomous replanning. Our proof-of-concept experiments show that mid-trajectory replanning substantially improves the success rate, but this capability currently relies on manual triggers. We believe that discovering a principled, autonomous trigger is a promising direction that would unlock GAC's full potential, likely closing the performance gap in complex tasks where open-loop plans are insufficient.



%% file: scripts/discussion.tex
This paper introduced the Generative Actor-Critic (GAC), a framework that shifts the paradigm of reinforcement learning from estimating expected returns to modeling the complete joint distribution of trajectories and their outcomes, $p(\tau,y)$. Our experiments validate this approach, showing that GAC not only achieves state-of-the-art performance but does so by learning remarkably structured internal representations. We argue that the generative objective itself—the need to explain the entire data distribution under the bottleneck of the latents—is what compels the model to move beyond observational pattern matching and form an implicit world model of its environment. The emergence of these properties from a general-purpose learning objective provides a powerful explanation for GAC's strong planning capabilities and suggests that modeling the full distribution might be a key step towards developing agents with a deeper, more fundamental understanding of their world.

The GAC framework orchestrates and extends concepts from several modern RL paradigms. In contrast to Distributional RL, its focus on entire trajectories enables holistic, long-horizon planning. Compared to sequence modeling approaches like Decision Transformer, whose decison-making rely on conditioning with a fixed, potentially \textit{biased} target, GAC offers a more principled approach through its distinct inference strategies for exploitation (GAC-$\mathbb{E}[y]$) and exploration (GAC-$p(y^+)$). While effective, a key limitation of our current approach is the manual tuning of the exploration increment $\Delta y$. Future work should focus on developing adaptive mechanisms to automate this process.

%% file: scripts/supp_related.tex
\textbf{Offline RL} algorithms train agents solely from pre-collected datasets\cite{levine2020offline}. Most work focuses on reducing extrapolation error from out-of-distribution (OOD) actions. Pessimistic methods lower value estimates for OOD state-action pairs to discourage unreliable actions\cite{yu2020mopo, kumar2020conservative, kidambi2020morel, luo2025learning}, while conservative approaches constrain policies to stay close to the behavior policy\cite{wu2019behavior, peng2019advantage, fujimoto2021minimalist, fujimoto2019off, kumar2019stabilizing}. Compared to develop methods on RL and Bellman equation, our method is more closed to generative modeling to learn a latent plan-conditioned policy~\citep{emmons2021rvs, rosete2023latent}.

\textbf{Exploration} is a major challenge in online RL. Exploration algorithms can be broadly categorized into two groups. For augmented data collection strategies, exploration can be encouraged by action selection perturbations~\citep{painter2023monte, zhou2020neural, wang2022thompson}, guided state selection~\citep{ecoffet2019go}, and parameter perturbations~\citep{fortunato2017noisy}. For augmented training strategies, techniques include count-based rewards~\citep{tang2017exploration, bellemare2016unifying, fu2017ex2}, prediction-based rewards~\citep{burda2018exploration, badia2020never, badia2020agent57}, entropy-augmented rewards~\citep{haarnoja2018soft}, and information-theoretic objectives~\citep{eysenbach2018diversity, houthooft2016vime}. In the LHBG model, exploration and exploitation are balanced by tuning the coefficients of value prediction and KL regularization.

\textbf{Offline-to-Online RL} methods aim to bridge the gap between the high exploration cost in online RL and the often suboptimal results of offline RL~\citep{luo2023finetuning}. Most existing approaches focus on mitigating the over-conservatism introduced by offline RL algorithms, such as through policy constraints~\citep{kostrikov2021offline, nair2020awac, li2023proto, beeson2022improving}, policy expansion~\citep{zhang2023policy}, value calibration~\citep{nakamoto2023cal}, or value ensembles~\citep{lee2022offline}. Some methods introduce generative models to generate data in a more flexible manner to address distribution shift~\citep{liu2024energy}. Unlike these approaches, our model addresses offline-to-online RL within a generative sequential framework and enables exploratory data collection and online fine-tuning with versatile inference processes after offline pretraining.

\textbf{Generative Modeling for Deicision-Making} has emerged as a new paradigm following the introduction of models such as Decision Transformer~\citep{chen2021decision} and Trajectory Transformer~\citep{janner2021offline}. After autoregressive models, diffusion have become influential alternatives for policy learning~\citep{chi2023diffusion} and planning~\citep{janner2022planning}, with flow matching also gaining attention~\citep{lipman2022flow, zheng2023guided}. Similar to our method, several latent variable models have been proposed for decision-making. For example, \citet{paster2022you, yang2022dichotomy} and \citet{huang2023reparameterized} introduce latent variables to address environmental stochasticity and action multi-modality, respectively. Our approach is closely related to and inspired by \citet{kong2024latent}, but is unique in its use of an ELBO-based variational Bayes learning and in formalizing various inference objectives especially the one for online data collection. Furthermore, our method differs in specific model architecture and data specification.

\textbf{Cross-Entropy Method} (CEM)~\citep{rubinstein1999cross} is a classical example of elite-based sampling methods. It is an iterative optimization algorithm that samples a population of solutions from a parameterized distribution, evaluates them, and then refits the distribution's parameters to the "elite" top-performing subset. This process progressively focuses the search on high-reward regions. While GAC does not implement a standard CEM, its exploration strategy GAC-$p(y^+)$ shares this core principle of iterative, elite-driven refinement. GAC-$p(y^+)$ treats its replay buffer as an empirical distribution of past returns and samples an ``elite'' set by drawing from its high-performing quantile. Instead of refitting a simple distribution, GAC performs posterior inference to sample latent plans conditioned on optimistic targets derived from this elite set. The resulting high-quality trajectories are added back to the buffer, gradually shifting the collected data distribution toward superior outcomes, effectively achieving a similar goal of iterative population improvement.

\textbf{Limitations of step-wise Markov rewards} have been recently discussed by \citet{abel2021expressivity} and further explored by \citet{bowling2023settling} and \citet{qin2023learning}: there are restrictions on what kinds of preferences over policies can be codified in terms of a step-wise reward function that is Markov. In contrast, in the framework of GAC, there is only a final return for a trajectory. This was exemplified by \citet{chen2021decision} and adopted by \citet{kong2024latent}. GAC is the first method that demonstrate performance competitive to methods with step-wise rewards in both offline and online learning. 

%% file: scripts/supp_exp.tex
\begin{algorithm}[H]
\caption{GAC Offline Pre-training}
\label{alg:offline}
\begin{algorithmic}[1]
\STATE \textbf{Input:} Offline dataset $\mathcal{D}$, model $\theta$, variational params $\phi$.
\STATE Initialize $\theta$ and $\{\phi_i\}_{i=1}^N$ for all data points.
\STATE \textbf{repeat}
\STATE \quad Sample batch $\mathcal{B} = \{(\tau_j, y_j, s_{j,0})\}$ from $\mathcal{D}$.
\STATE \quad \textit{// Inner Loop: Local Variational Inference}
\STATE \quad \textbf{for} each $j \in \mathcal{B}$ \textbf{do}
\STATE \quad \quad Initialize $\phi_j$. Maximize ELBO (\textbf{Eq.~\eqref{eq:elbo}}) w.r.t. $\phi_j$ until convergence to obtain $\phi_j^*$.
\STATE \quad \textbf{end for}
\STATE \quad \textit{// Outer Loop: Global Parameter Learning}
\STATE \quad Compute $\nabla_\theta \text{ELBO}$ using optimized posteriors $q_{\phi^*}$.
\STATE \quad Update $\theta \leftarrow \theta + \eta_{\theta} \nabla_{\theta} \text{ELBO}$.
\STATE \textbf{until} convergence of $\theta$.
\STATE \textbf{Return:} Pre-trained parameters $\theta$.
\end{algorithmic}
\end{algorithm}

\begin{algorithm}[H]
\caption{GAC Online Fine-tuning}
\label{alg:online}
\begin{algorithmic}[1]
\STATE \textbf{Input:} Pre-trained $\theta$, Env, Buffer $\mathcal{R} \leftarrow \mathcal{D}$, Stages $S$, steps $C, F$, $\Delta y$, top-$k$.
\STATE \textbf{for} stage $s = 1 \dots S$ \textbf{do}
\STATE \quad \textit{// --- Data Collection (Exploration) ---}
\STATE \quad \textbf{for} $c = 1 \dots C$ \textbf{do}
\STATE \quad \quad $s_0 \leftarrow Env$. Sample base $y$ from top-$k$ of $\mathcal{R}$. Set $y^+ = y + \Delta y$.
\STATE \quad \quad Infer plan $z \sim q_\phi^*(z)$ via \textbf{Exploration Query} (see Alg.~\ref{alg:inference}) targeting $y^+$.
\STATE \quad \quad Generate $\tau \sim p_\beta(\cdot|z, s_0)$, execute, observe $y_{true}$, update $\mathcal{R} \leftarrow \mathcal{R} \cup \{(\tau, y_{true})\}$.
\STATE \quad \textbf{end for}
\STATE \quad \textit{// --- Model Fine-tuning ---}
\STATE \quad \textbf{for} $f = 1 \dots F$ \textbf{do}
\STATE \quad \quad Update $\theta$ on batch from $\mathcal{R}$ maximizing ELBO (as in Alg.~\ref{alg:offline}).
\STATE \quad \textbf{end for}
\STATE \textbf{end for}
\end{algorithmic}
\end{algorithm}

\begin{algorithm}[H]
\caption{GAC Inference Strategies (Test-time \& Exploration)}
\label{alg:inference}
\begin{algorithmic}[1]
\STATE \textbf{Input:} Query Type, initial state $s_0$, model components $p_\alpha, p_\beta, p_\gamma$.
\STATE \textbf{Switch} Query Type:
\STATE \quad \textbf{Case Exploitation ($\max \mathbb{E}[y]$):}
\STATE \quad \quad \textit{// Active optimization for best expected return}
\STATE \quad \quad Find $z^* = \arg\max_z \mathbb{E}_{p_\gamma}[y|z]$ via gradient ascent, initialized from prior $p_\alpha(z|s_0)$.
\STATE \quad \quad \textbf{Return} $z^*$.
\STATE \quad \textbf{Case Exploration (Sampling from $p(z|y^+)$):}
\STATE \quad \quad \textit{// Principled posterior sampling with optimistic target}
\STATE \quad \quad Given target $y^+$, solve for $q_\phi^*(z)$ maximizing $\mathbb{E}_q[\log p_\gamma(y^+|z)] - D_{KL}(q\|p_\alpha)$.
\STATE \quad \quad \textbf{Return} $z \sim q_\phi^*(z)$.
\STATE \quad \textbf{Case Conditional ($y=y^*$):}
\STATE \quad \quad \textit{// Standard "Decision Transformer or Diffuser" style conditioning}
\STATE \quad \quad Similar to Exploration, but target $y^*$ is fixed (e.g., max possible return).
\STATE \quad \quad \textbf{Return} $z \sim q_{\phi, y^*}^*(z)$.
\STATE \quad \textbf{Case Prior Sampling:}
\STATE \quad \quad \textit{// Diverse generation without target guidance}
\STATE \quad \quad \textbf{Return} $z \sim p_\alpha(z|s_0)$.
\STATE \textbf{End Switch}
\STATE Generate trajectory $\tau \sim p_\beta(\tau|z, s_0)$ using the returned $z$.
\end{algorithmic}
\end{algorithm}
\vspace{20pt}

\begin{table}[h]
    \centering
    \caption{Comparison between Latent Plan Transformer (LPT) and Generative Actor-Critic (GAC).}
    \small
    \label{tab:gac_vs_lpt}
    \begin{tabular}{l|p{5.5cm}|p{5.5cm}}
    \toprule
    \textbf{Aspect} & \textbf{LPT} & \textbf{GAC (Ours)} \\
    \midrule
    \textbf{Latent $z$} & Monolithic $z$ generates full $\tau$ (including $s_0$). & Decoupled $z_y, z_{\setminus y}$. Prior conditioned on $s_0$. \\
    \midrule
    \textbf{Initial State $s_0$} & Generated ($s_0 \sim p(\tau|z)$); risks mismatch. & Conditioned (Given $s_0$); anchors plan to reality. \\
    \midrule
    \textbf{Exploitation} & \textbf{Conditional Sampling}: $p(z|y=y^*)$ with fixed target. & \textbf{Active Optimization}: Gradient ascent on $\mathbb{E}[y|z]$. \\
    \midrule
    \textbf{Exploration} & Stochastic sampling (heuristic); no dedicated mechanism. & \textbf{Posterior Sampling}: $z \sim p(z|y^+)$ with dynamic targets. \\
    \midrule
    \textbf{Inference} & \textbf{MCMC}: Slow convergence, high cost. & \textbf{Gradient Ascent}: Fast, stable, lower loss. \\
    \midrule
    \textbf{Paradigm} & Primarily Offline RL. & Active Online RL (Explore-Collect-Train). \\
    \bottomrule
    \end{tabular}
\end{table}

\begin{figure}[t]
    \centering
    \includegraphics[width=1.0\textwidth]{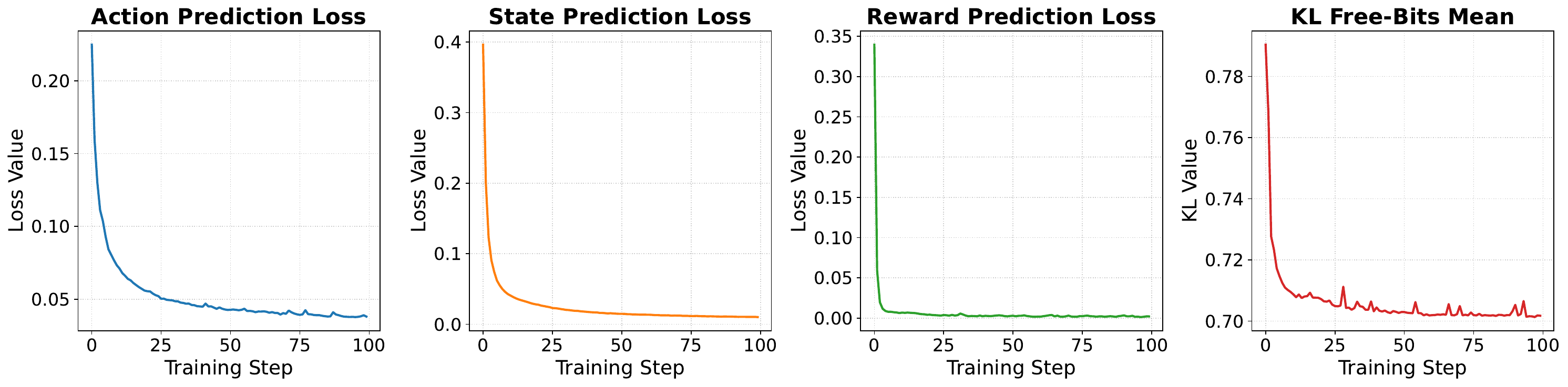} 
    \caption{\textbf{Training curve of hopper-medium dataset.} We visualize the evolution of the four components of the ELBO objective during offline pre-training. The reconstruction losses for actions, states, and returns exhibit consistent monotonic convergence, indicating that the model effectively learns to approximate the joint distribution $p(\tau, y)$ via the latent bottleneck. Originating from a non-zero value due to the standard normal initialization of variational parameters, the KL term gradually decreases and stabilizes. This confirms that the softplus-based free-bits mechanism successfully regularizes the latent space, preventing posterior collapse while ensuring smooth gradient updates.}
    \label{fig:training_curve}
\end{figure}

We employ a more sophisticated alternate training methodology than that of LPT \citep{kong2024latent}. Specifically, for each data point, we initialize the variational parameters (mean and log-variance) by sampling from a standard normal distribution, and then conduct inner-loop optimization until convergence, which is defined by a maximum of 100 training steps and an early-stopping threshold of 1e-4. This random initialization prevents the optimization from starting at a trivial zero-KL state; consequently, the KL divergence typically initiates at a non-zero value and exhibits a gradual descent as the posterior is refined. Furthermore, we implement the \textit{free bits} mechanism \cite{kingma2016improved} with a softplus transition $\eta + \text{softplus}(D_{KL} - \eta)$ with free bits margin $\eta$ to ensure smooth gradients. In the outer loop, unlike the single-step optimization in LPT, we treat the number of optimization steps as a tunable hyperparameter. This approach aims to tighten the approximation between the Evidence Lower Bound (ELBO) and the log-likelihood, and mitigating the issue of posterior collapse described as \cite{li2023overcoming}. This rigorous bi-level optimization scheme demonstrates high stability; as illustrated in \cref{fig:training_curve} , the ELBO converges smoothly and monotonically during offline pre-training, confirming the effectiveness of our alternating update strategy.

Given that GAC decouples decision-modeling from decision-making, we can directly assess the quality of our generative model by its loss value, rather than relying on immediate policy evaluation in the environment. Consequently, we leverage the Optuna framework \cite{akiba2019optuna} for hyperparameter optimization, with the objective of maximizing the ELBO. The complete set of hyperparameters subject to tuning is detailed in \cref{tab:hyperparams}. Following the pre-training phase, architecture-related hyperparameters are fixed, while the remaining parameters are optimized during each subsequent fine-tuning stage. Using the optimal hyperparameters identified (see \cref{tab:best_arch_hyperparams} and \cref{tab:best_other_hyperparams}\footnote{The hyperparameters with interval range are fp64 after searching, only displayed four decimal places in \cref{tab:best_other_hyperparams}. 'h', 'w', 'ha', 'm', 'r', 'u' and 'l' denotes 'hopper', 'walker2d', 'halfcheetah', 'medium', 'replay', 'umaze' and 'large' respectively.}), we then configure the inner-loop inference with more stringent convergence criteria: a maximum of 1000 steps and an early-stopping threshold of 1e-5. The offline pre-training and online fine-tuning algorithms are described by \cref{alg:offline} and \cref{alg:online} in the appendix.

\begin{table}[t]
\vspace{-20pt}
\caption{GAC Hyperparameters.}
\begin{center}
\begin{tabular}{cc}
\toprule
Architecture-Related & Range \\
\midrule
Embedding dimension & choice(64, 128, 192, 256, 512) \\
Number of latent tokens & choice(1, 2, 4, 8) \\
Number of attention heads & choice(1, 2, 4, 8) \\
Number of decoder layers & choice(1, 2, 3, 4) \\
Number of encoder layers & choice(1, 2, 3, 4) \\
Context length & choice(1, 4, 16, 32, 64) \\
\midrule
Architecture-Unrelated & Range \\
\midrule
Outer-loop learning rate & interval(1e-4, 1e-2) \\
Outer-loop weight decay & interval(1e-4, 1e-2) \\
Outer-loop training steps & choice(1, 5, 10, 20) \\
Inner-loop learning rate & interval(0.01, 0.5) \\
Batch size & choice(800, 700, 600, 500) \\

\bottomrule
\end{tabular}
\end{center}
\label{tab:hyperparams}
\end{table}

\begin{table}[t]
\caption{The best GAC Architecture-Related Hyperparameters.}
\begin{center}
\begin{small}
\begin{tabular}{c|ccccccccc}
\toprule
Parameter & h-m & h-m-r & w-m & w-m-r & ha-m & ha-m-r & m-u & m-m & m-l \\
\midrule
Embedding dimension & 192 & 128 & 192 & 192 & 128 & 256 & 192 & 256 & 192 \\
Number of latent tokens & 4 & 8 & 4 & 4 & 8 & 4 & 4 & 8 & 2 \\
Number of attention heads & 4 & 4 & 2 & 8 & 1 & 4 & 1 & 2 & 2 \\
Number of decoder layers & 3 & 4 & 3 & 4 & 3 & 3 & 2 & 2 & 2 \\
Number of encoder layers & 4 & 1 & 3 & 3 & 2 & 2 & 1 & 3 & 2 \\
Context length & 4 & 4 & 1 & 1 & 4 & 1 & 4 & 16 & 4 \\
\bottomrule
\end{tabular}
\end{small}
\end{center}
\label{tab:best_arch_hyperparams}
\vspace{-10pt}
\end{table}

\begin{table}[t]
\caption{The best GAC Architecture-Unrelated Hyperparameters.}
\begin{center}
\begin{smaller}
\begin{tabular}{c|ccccccccc}
\toprule
Parameter & h-m & h-m-r & w-m & w-m-r & ha-m & ha-m-r & m-u & m-m & m-l \\
\midrule
\multicolumn{10}{c}{Stage 0 (pre-training)} \\
\midrule
Outer-loop learning rate & 0.0010 & 0.0003 & 0.0023 & 0.0026 & 0.0026 & 0.0012 & 0.0025 & 0.0002 & 0.0009 \\
Outer-loop weight decay & 0.0002 & 0.0056 & 0.0027 & 0.0011 & 0.0020 & 0.0071 & 0.0005 & 0.0002 & 0.0015 \\
Outer-loop training steps & 10 & 1 & 5 & 1 & 10 & 10 & 5 & 1 & 20 \\
Inner-loop learning rate & 0.2365 & 0.2058 & 0.3980 & 0.0795 & 0.0501 & 0.0452 & 0.0142 & 0.08777 & 0.1172 \\
Batch size & 700 & 500 & 600 & 500 & 500 & 700 & 600 & 500 & 500 \\
\midrule
\multicolumn{10}{c}{Stage 1 (fine-tuning)} \\
\midrule
Outer-loop learning rate & 0.0002 & 0.0018 & 0.0008 & 0.0017 & 0.0062 & 0.0009 & 0.0013 & 0.0034 & 0.0010  \\
Outer-loop weight decay & 0.0002 & 0.0026 & 0.0029 & 0.0062 & 0.0044 & 0.0060 & 0.0008 & 0.0003 & 0.0007 \\
Outer-loop training steps & 1 & 20 & 1 & 20 & 5 & 20 & 1 & 1 & 1 \\
Inner-loop learning rate & 0.0458 & 0.2116 & 0.1223 & 0.0712 & 0.2363 & 0.2836 & 0.0573 & 0.1163 & 0.0712 \\
Batch size  & 800 & 700 & 800 & 500 & 700 & 500 & 600 & 500 & 500 \\
\midrule
\multicolumn{10}{c}{Stage 2 (fine-tuning)} \\
\midrule
Outer-loop learning rate & 0.0004  & 0.0010 & 0.0013 & 0.0014 & 0.0093 & 0.0013 & 0.0016 & 0.0028 & 0.0077 \\
Outer-loop weight decay & 0.0006 & 0.0007 & 0.0022 & 0.0050 & 0.0001 & 0.0002 & 0.0006 & 0.0017 & 0.0001 \\
Outer-loop training steps & 1 & 1 & 5 & 20 & 5 & 5 & 1 & 1 & 10 \\
Inner-loop learning rate & 0.1018 & 0.0712 & 0.0124 & 0.1691 & 0.1435 & 0.2154 & 0.2508 & 0.0316 & 0.2089 \\
Batch size  & 700 & 500 & 800 & 500 & 600 & 700 & 500 & 800 & 800 \\
\midrule
\multicolumn{10}{c}{Stage 3 (fine-tuning)} \\
\midrule
Outer-loop learning rate & 0.3755 & 0.0005 & 0.0007 & 0.0010 & 0.0039 & 0.0022 & 0.0032 & 0.0017 & 0.0005 \\
Outer-loop weight decay & 0.0077 & 0.0001 & 0.0027 & 0.0007 & 0.0023 & 0.0015 & 0.0030 & 0.0066 & 0.0024 \\
Outer-loop training steps & 5 & 5 & 20 & 1 & 20 & 20 & 1 & 5 & 20 \\
Inner-loop learning rate & 0.0023 & 0.010 & 0.0337 & 0.0712 & 0.1193 & 0.1984 & 0.1030 & 0.2807 & 0.1491 \\
Batch size  & 600 & 500 & 600 & 500 & 500 & 600 & 700 & 700 & 500 \\
\bottomrule
\end{tabular}
\end{smaller}
\end{center}
\label{tab:best_other_hyperparams}
\end{table}

For GAC-$p(y^+)$ to collect online data, we introduce another two parameters to obtain a target return set. We perform a data processing procedure on the current return dataset, which we denote as $D$. This process involves a quantile-based filtering and a subsequent transformation. The procedure is parameterized by two tunable hyperparameters: the quantile threshold, $q$, and an additive constant, $\Delta y$. First, we determine the $q$-th quantile of the dataset $D$, which we denote as $y_q$. All data points in $D$ that are greater than $y_q$ are selected to form a new subset, $D_\text{sub}=\{y\in D \mid y>y_q\}$. We finally randomly sample from $D_\text{sub}$ and plus $\Delta y$ to obtain the dynamic target return as $\{y+\Delta y \mid y\in D_\text{sub}\}$. This procedure allows us to isolate and transform the upper tail of the data distribution, with the specific threshold and transformation intensity controlled by the hyperparameters $q$ and $\Delta y$, respectively. These additional parameters for inference and the standard target return for each environment in \cref{tab:inferece_params}. We present a more complete offline and online results in \cref{tab:supp_offline_results} and \cref{tab:supp_online_results}, containing outcomes from both step-wise rewards and final return only. 

\begin{table}[t]
\caption{The Inference Hyperparameters.}
\begin{center}
\begin{smaller}
\begin{tabular}{c|ccccccccc}
\toprule
Parameter & h-m & h-m-r & w-m & w-m-r & ha-m & ha-m-r & m-u & m-m & m-l \\
\midrule
Standard target return & 3600 & 3600 & 6000 & 6000 & 5000 & 5000 & 165 & 280 & 275 \\
Stage0 $(q,\Delta y)$ & (0.8,10) & (0.8,10) & (0.8,10) & (0.8,10) & (0.6,3) & (0.6,3) & (0.8, 100) & (0.8, 100) & (0.8, 100) \\
Stage1 $(q,\Delta y)$ & (0.8,10) & (0.8,10) & (0.8,10) & (0.8,10) & (0.6,3) & (0.6,3) & (0.8, 100) & (0.8, 100) & (0.8, 100) \\
Stage2 $(q,\Delta y)$ & (0.8,10) & (0.8,10) & (0.6,3) & (0.6,5) & (0.6,3) & (0.6,1) & (0.8, 100) & (0.8, 100) & (0.8, 100) \\
Stage3 $(q,\Delta y)$ & (0.8,10) & (0.8,10) & (0.6,3) & (0.6,5) & (0.6,3) & (0.6,1) & (0.8, 100) & (0.8, 100) & (0.8, 100) \\
\bottomrule
\end{tabular}
\end{smaller}
\end{center}
\label{tab:inferece_params}
\end{table}

\begin{table}[t]
\caption{\textbf{Offline pre-training results for MuJoCo and Maze2D.} Comparison between the average normalized returns of GAC against several baselines, evaluated without any online finetuning. We assess GAC with the exploitation query, denoted as GAC-$\mathbb{E}[y]$. For each of 5 random seeds, we generate 100 evaluation trajectories and report the mean and standard deviation of their returns. The best result in each row is highlighted in bold.}
\begin{center}
\begin{smaller}
\setlength{\tabcolsep}{4pt}
\begin{tabularx}{\textwidth}{@{}l|CCCC|CCCCCcc@{}}
\toprule
\multirow{2}{*}{Dataset} & \multicolumn{4}{c|}{Step-wise Reward} & \multicolumn{7}{c}{Final Return} \\
\cmidrule(lr){2-5} \cmidrule(lr){6-12}
& IQL & CQL & DT & QDT & IQL & CQL & DT & QDT & LPT & GAC-$\mathbb{E}[y]$ & GAC-$p(y^+)$ \\
\midrule
hopper-medium & 63.8 & 58.0 & 60.3 & 66.5 & 35.1 & 23.3 & 57.3 & 50.7 & 58.5 & 60.2${\pm}$8.0 & \textbf{73.2}${\pm}$13.6 \\
hopper-medium-replay & 92.1 & 48.6 & 63.7 & 52.1 & 13.9 & 7.7 & 50.8 & 38.7 & 71.2 & \textbf{80.1}${\pm}$9.2 & 67.5${\pm}$25.1 \\
walker2d-medium & 79.9 & 79.2 & 73.3 & 67.1 & 49.1 & 0.0 & 69.9 & 63.7 & 77.8 & \textbf{80.2}${\pm}$4.6 & 78.9${\pm}$8.9 \\
walker2d-medium-replay & 73.7 & 74.1 & 60.2 & 58.2 & 5.3 & 3.2 & 51.6 & 29.6 & 72.3 & \textbf{78.9}${\pm}$9.3 & 73.3${\pm}$10.2 \\
halfcheetah-medium & 47.4 & 44.4 & 42.1 & 42.3 & 8.5 & 1.0 & 42.4 & 42.4 & 43.1 & \textbf{43.6}${\pm}$2.5 & 41.5${\pm}$5.1 \\
halfcheetah-medium-replay & 44.1 & 46.2 & 34.1 & 35.6 & 5.2 & 7.8 & 33.3 & 32.8 & 39.6 & \textbf{39.8}${\pm}$7.8 & 38.8${\pm}$8.9 \\
\midrule
maze2d-umaze & 37.7 & 5.7 & 31.0 & 57.3 & 4.5 & 3.9 & 28.4 & 2.57 & 65.4 & \textbf{67.8}${\pm}$21.4 &  64.2${\pm}$22.7 \\
maze2d-medium & 35.5 & 5.0 & 8.2 & 13.3 & 3.5 & -3.6 & -2.4 & -2.58 & 20.6 & \textbf{74.5}${\pm}$71.0 & 63.3${\pm}$71.4 \\
maze2d-large & 49.6 & 12.5 & 2.3 & 31.0 & 2.0 & -1.2 & -2.5 & -2.51 & 37.2 & \textbf{50.3}${\pm}$40.4 &  39.5${\pm}$42.3 \\
\bottomrule
\end{tabularx}
\end{smaller}
\end{center}
\label{tab:supp_offline_results}
\end{table}

\begin{table}[t]
\vspace{-5pt}
\caption{\textbf{Online fine-tuning results.} Comparison of normalized returns before and after online fine-tuning with only access to final return. We fine-tune GAC for 3 stages with data collected by GAC-$p(y^+)$ and report the final GAC-$\mathbb{E}[y]$ performance from five seeds. The best final result for each dataset is highlighted in bold.}
\label{tab:supp_online_results}
\centering
\smaller
\setlength{\tabcolsep}{4pt}
\begin{tabularx}{\textwidth}{l|c|c|Cc|cc|Cc|Cc|Cc|cc}
\toprule
Dataset
& \multicolumn{2}{c|}{Online RL}
& \multicolumn{2}{c|}{ODT}
& \multicolumn{2}{c|}{CQL}
& \multicolumn{2}{c|}{Cal-QL}
& \multicolumn{2}{c|}{IQL}
& \multicolumn{2}{c|}{LPT}
& \multicolumn{2}{c}{GAC} \\
& PPO & SAC & online & $\delta$ & online & $\delta$ & online & $\delta$ & online & $\delta$ & online & $\delta$ & online & $\delta$ \\
\midrule
hopper-m & \multirow{2}{*}{13.1${\pm}$2.1} & \multirow{2}{*}{11.2${\pm}$1.5} & 57.6 & +0.3 & 29.6 & +6.3 & 32.8 & +9.5 & 25.0 & -10.1 & 64.8 & +6.3 & \textbf{93.3}${\pm}$5.1 & +33.1 \\
hopper-m-r &  &  & 65.2 & +14.4 & 8.4 & +0.7 & 24.1 & +16.4 & 12.6 & -1.3 & 72.4 & +1.2 & \textbf{95.2}${\pm}$1.9 & +15.1 \\
walker2d-m & \multirow{2}{*}{9.5${\pm}$1.6} & \multirow{2}{*}{4.0${\pm}$0.9} & 70.7 & +0.8 & 1.9 & +1.9 & 1.2 & +1.2 & 50.1 & +1.0 & 79.5 & +1.7 & \textbf{85.1}${\pm}$4.9 & +4.9 \\
walker2d-m-r &  &  & 57.3 & +5.7 & 0.5 & -2.7 & 3.5 & +0.3 & 6.9 & +1.6 & 79.0 & +6.7 & \textbf{85.4}${\pm}$5.3 & +6.5 \\
halfcheetah-m & \multirow{2}{*}{1.3${\pm}$0.1} & \multirow{2}{*}{1.7${\pm}$0.2} & 40.7 & -1.7 & 2.8 & +1.8 & 3.1 & +2.1 & 8.9 & +0.4 & 43.2 & +0.1 & \textbf{44.3}${\pm}$1.1 & +0.8 \\
halfcheetah-m-r &  &  & 24.4 & -8.4 & 6.4 & -1.4 & 2.3 & -5.5 & 7.4 & +2.2 & 40.6 & +1.0 & \textbf{40.9}${\pm}$0.8 & +1.1 \\
\midrule
maze2d-umaze & 55.5${\pm}$1.1 & 61.8${\pm}$1.2 & 11.2 & -17.2 & 5.8 & +1.9 & 4.9 & +1.0 & 29.3 & +24.8 & 67.2 & +1.8 & \textbf{83.5}${\pm}$23.9 & +15.7 \\
maze2d-medium & 29.3${\pm}$3.9 & 46.8${\pm}$4.7 & 2.5 & +4.9 & -1.8 & +1.8 & 0.9 & +4.5 & 23.1 & +19.6 & 26.1 & +5.5 & \textbf{166.1}${\pm}$33.4 & +91.6 \\
maze2d-large & -0.8${\pm}$0.8 & 17.7${\pm}$0.8 & 2.2 & +4.7 & 0.3 & +1.5 & 1.1 & +2.3 & 7.4  & +5.4 & 40.1 & +2.9 & \textbf{94.2}${\pm}$42.7 & +43.9  \\
\midrule
\multicolumn{15}{c}{Step-wise Reward} \\
\midrule
hopper-m & \multirow{2}{*}{77.8${\pm}$1.3} & \multirow{2}{*}{75.0${\pm}$5.8} & 97.5 & +30.6 & 60.4 & +2.4 & 98.0 & +40.0 & 66.8 & +3.0 & - & - & - & - \\
hopper-m-r &  &  & 88.9 & +2.3 & 56.3 & +7.7 & 110.0 & +61.4 & 96.2 & +4.1 & - & - & - & - \\
walker2d-m & \multirow{2}{*}{43.9${\pm}$1.4} & \multirow{2}{*}{47.5${\pm}$1.2} & 76.8 & +4.6 & 79.6 & +0.4 & 103.0 & +23.8 & 80.3 & +0.4 & - & - & - & - \\
walker2d-m-r &  &  & 76.9 & +4.0 & 75.8 & +1.7 & 99.0 & +24.9 & 70.6 & -3.1 & - & - & - & -  \\
halfcheetah-m & \multirow{2}{*}{25.0${\pm}$2.6} & \multirow{2}{*}{24.4${\pm}$2.7} & 42.4 & -0.6 & 45.6 & +1.2 & 93.0 & +48.6 & 47.4 & +0.0 & - & - & - & -  \\
halfcheetah-m-r &  &  & 40.2 & +0.4 & 43.0 & -3.2 & 93.0 & +46.8 & 44.1 & +0.0 & - & - & - & -  \\
\midrule
maze2d-umaze & 119.7${\pm}$1.2 & 115.6${\pm}$1.3 & 14.0 & -17.0 & 6.0 & +0.3 & 7.4 & +1.7 & 98.6 & +60.9 & - & - & - & - \\
maze2d-medium & 153.4${\pm}$0.9 & 123.8${\pm}$1.0 & 4.9 & -3.3 & 8.2 & +3.2 & 8.4 & +3.4 & 62.2 & +26.7 & - & - & - & - \\
maze2d-large & 106.7${\pm}$9.6 & 96.4${\pm}$1.4 & 5.1 & +2.8 & 5.1 & -7.4 & 10.6 & -1.9 & 77.6 & +28.0 & - & - & - & - \\
\bottomrule
\end{tabularx}
\vspace{-10pt}
\end{table}

\begin{table}[t]
\caption{\textbf{Comparing inference objectives for pretrained GAC.} The mean and standard deviation of returns over 100 rollouts from exploitation query (GAC-$\mathbb{E}[y]$), exploration query (GAC-$p(y^+)$), fixed target steering (GAC-$y^*$), and sampling from the prior ($p(y|z)p(z)$).}
\begin{center}
\begin{small}
\begin{tabular}{@{}l|ccc|c}
\toprule
Dataset & GAC-$\mathbb{E}[y]$ & GAC-$p(y^+)$ & GAC-$y^*$ & $p(y|z)p(z)$ \\
\midrule
\multicolumn{5}{c}{Stage 0 (pre-training)} \\
\midrule
maze2d-umaze & 67.8${\pm}$21.4 & 64.2${\pm}$22.7 & 59.2${\pm}$21.6 & 35.7${\pm}$23.4 \\
maze2d-medium & 74.5${\pm}$71.0 & 63.3${\pm}$71.4 & 61.2${\pm}$70.2 & 19.9${\pm}$56.6 \\
maze2d-large & 50.3${\pm}$40.4 & 39.5${\pm}$42.3 & 28.9${\pm}$39.0 & -0.4${\pm}$6.1 \\
\midrule
\multicolumn{5}{c}{Stage 1 (fine-tuning)} \\
\midrule
maze2d-umaze & 70.7${\pm}$20.8 & 68.2${\pm}$21.2 & 62.9${\pm}$19.8 & 61.6${\pm}$19.6 \\
maze2d-medium & 91.4${\pm}$80.5 & 84.9${\pm}$81.8 & 100.6${\pm}$78.4 & 48.2${\pm}$84.6 \\
maze2d-large & 67.4${\pm}$40.5 & 36.3${\pm}$43.4 & 32.3${\pm}$40.5 & 37.6${\pm}$40.4 \\
\midrule
\multicolumn{5}{c}{Stage 2 (fine-tuning)} \\
\midrule
maze2d-umaze & 73.1${\pm}$20.1 & 72.3${\pm}$22.6 & 70.5${\pm}$22.1 & 63.0${\pm}$23.5 \\
maze2d-medium & 157.5${\pm}$47.4 & 122.7${\pm}$75.3 & 88.8${\pm}$80.7 & 84.8${\pm}$77.4 \\
maze2d-large & 71.7${\pm}$49.8 & 56.5${\pm}$56.3 & 55.2${\pm}$55.5 & 52.2${\pm}$56.0 \\
\midrule
\multicolumn{5}{c}{Stage 3 (fine-tuning)} \\
\midrule
maze2d-umaze & 83.5${\pm}$24.5 & 82.3${\pm}$24.8 & 72.4${\pm}$24.8 & 64.0${\pm}$28.0 \\
maze2d-medium & 166.1${\pm}$33.4 & 129.1${\pm}$78.3 & 75.9${\pm}$89.8 & 41.4${\pm}$69.0 \\
maze2d-large & 92.5${\pm}$44.1 & 91.8${\pm}$44.3 & 84.6${\pm}$48.4 & 67.6${\pm}$46.1 \\
\bottomrule
\end{tabular}
\end{small}
\end{center}
\label{tab:supp_inference}
\end{table}

\begin{figure}[h]
    \centering
    \includegraphics[width=0.65\textwidth]{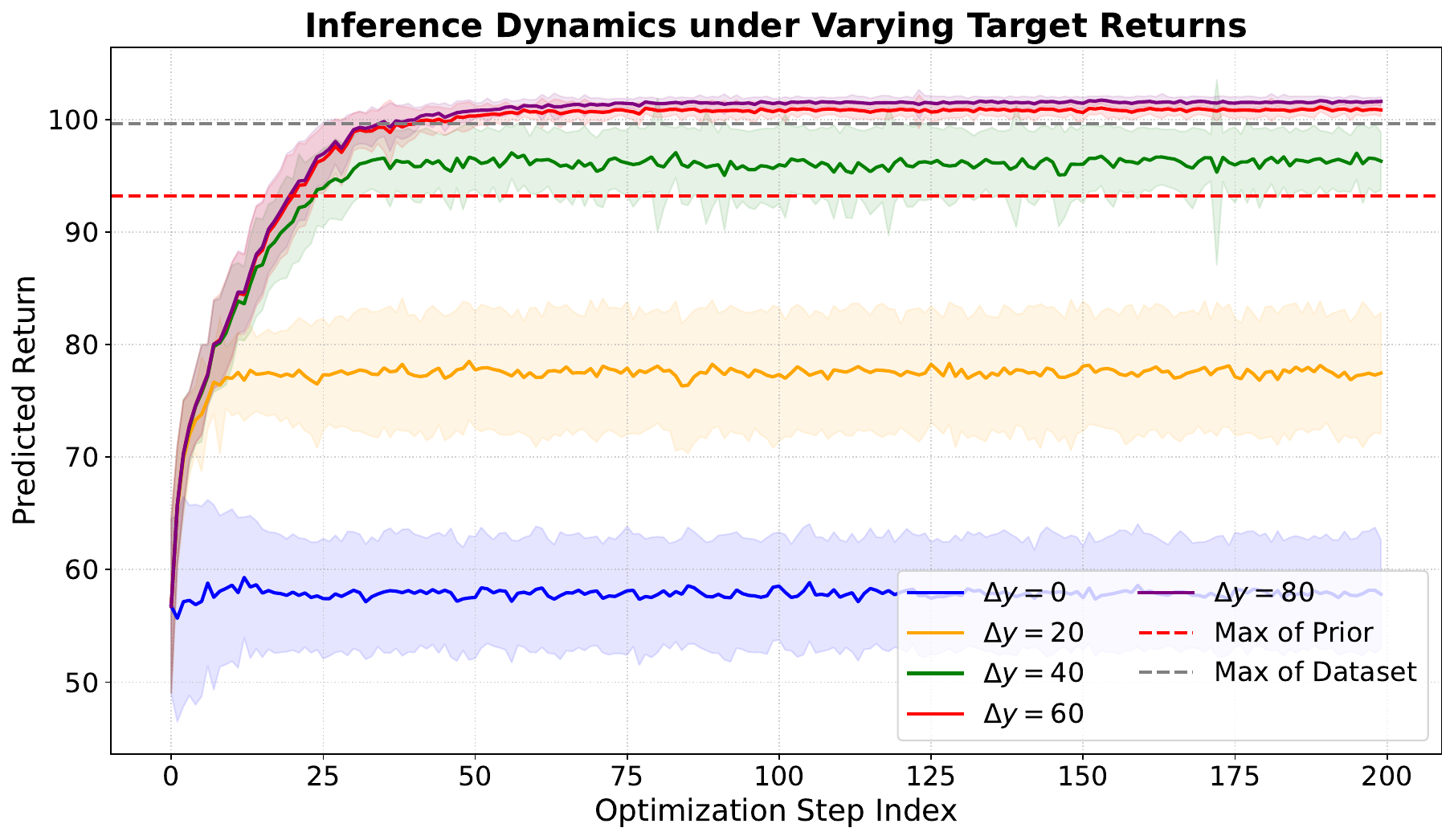} 
    \caption{\textbf{Inference dynamics under varying target returns.} We visualize the optimization trajectories of the latent plan's predicted return (mean $\pm$ std) over 200 steps with different $\Delta y$ in hopper-medium dataset. The red dashed line marks the maximum return of prior prediction.}
    \label{fig:sniper}
\end{figure}

%% file: scripts/supp_latent.tex
\subsection{Quantitative Validation of Latent Topology}

To address the concern regarding the learned latent structure shown in \cref{fig:online_results_maze} being merely (1) an artifact of dimensionality reduction (t-SNE) and (2) a trivial Lipschitz continuous mapping from the initial state $s_0$, we provide a comprehensive quantitative analysis performed in the original latent space, alongside with additional visualization. 

\begin{figure}[h]
    \centering
    \includegraphics[width=0.8\textwidth]{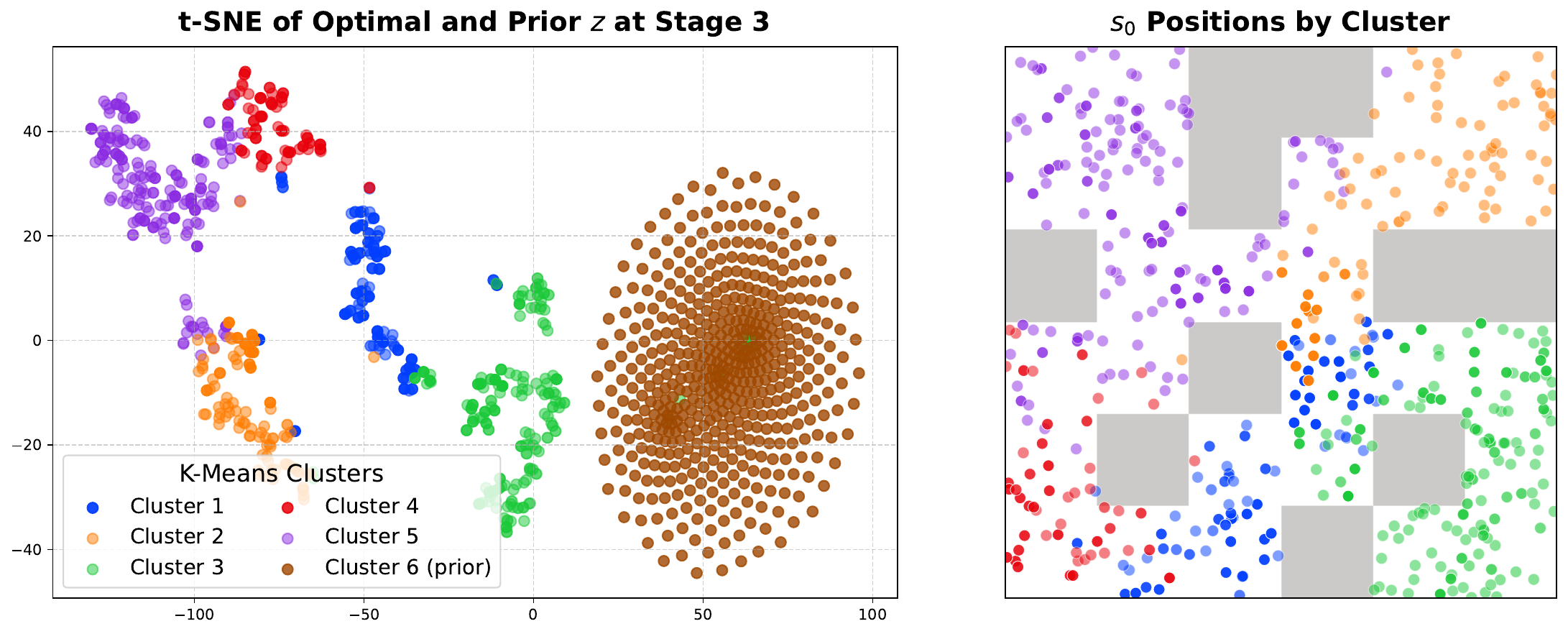}
    \caption{\textbf{Visualization of prior and optimized latent plans.} Prior samples form Cluster 6 (Brown), which is clearly separated from optimized latent plans (Clusters 1-5), demonstrating that latent optimization actively drives the latent plans away from the prior distribution to form task-specific clusters.}
    \label{fig:latent_structure}
\end{figure}

\textbf{Re-cluster with prior plans.} The clusters visualized in \cref{fig:online_results_maze} are optimal $z^*$ from $\arg\max_z \mathbb{E}[y|z]$. To show that the optimization, rather than the initial state $s_0$, is the key factor underlying the clustering, we mix the prior and the optimal plans and redo clustering. The new result is visualized in \cref{fig:latent_structure} where we color the $z^*$ clusters mirroring \cref{fig:online_results_maze}, and color the additional cluster from $z_{\text{prior}}$ in Brown. 

\textbf{Optimization shift.}
Because the global distance in t-SNE plots are not reliable, we also measure the Euclidean distance between the optimal plan $z^*$ and the initial sample from the prior $z_{\text{prior}} \sim p(z|s_0)$. As shown in \cref{tab:latent_metrics}, the average shift is $42.87$, which is substantially larger than the average intra-cluster distance $20.53$. 

\textbf{Cluster validity.}
Table~\ref{tab:latent_metrics} also compares the average inter-centroid distance and intra-cluster distance, with the former consistently larger than the latter. This confirms that the clusters observed in the t-SNE in \cref{fig:latent_structure} are well separated.

\textbf{Pair-wise inter-cluster distances.} 
We further provide the full pairwise distance matrix between cluster centroids in \cref{tab:dist_matrix}. Cluster 3 (Green) and Cluster 2 (Orange) have the maximal distance ($53.21$) among all cluster pairs, even though their initial states are geometrically adjacent. This separation occurs because the wall between the green and the orange regions warps the task-specific topology away from Euclidean. The model has learned a task-relevant topological representation based on navigational affordance.

\subsection{Inference Dynamics under OOD Targets}
\label{app:ood_inference}

To investigate the robustness of GAC's posterior inference $p(z|y)$, we track the posterior optimization dynamics of latent plans under varying degrees of Out-of-Distribution (OOD) targets. Figure \ref{fig:sniper} visualizes the predicted returns over 200 optimization steps for target increments $\Delta y$ ranging from $0$ to $80$. Two key properties emerge.

\textbf{Bounded Optimism via KL Anchoring.} As $\Delta y$ increases, the inferred returns do not diverge linearly. Instead, the trajectories for $\Delta y \ge 60$ (red, purple, brown lines) rapidly converge to a common asymptote, only slightly exceeding the prior maximum. This confirms that the KL-divergence term in the ELBO acts as a robust ``soft bound,'' anchoring the posterior to the feasible manifold of the prior and preventing the hallucination of unrealistic returns.

\textbf{OOD Snipering (Variance Collapse).} A clear transition in uncertainty is observed. Low targets (e.g., $\Delta y=0$, blue) maintain high variance, reflecting the diversity of sub-optimal behavior. In contrast, optimistic targets trigger a variance collapse, where the probability mass concentrates on a narrow, sparse region of optimal plans. This demonstrates that GAC effectively ``snipers'' the optimal solution manifold when driven by high-value OOD targets, with minimum risk of overshoot.

\begin{table}[h]
    \centering
    \small
    \caption{\textbf{Euclidean distances in latent space.}}
    \label{tab:latent_metrics}
    \begin{tabular}{lcc}
        \toprule
        Metric & Mean Value & Std. Dev \\
        \midrule
        Optimization Shift ($||z^* - z_{\text{prior}}||_2$) & 42.87 & 17.50 \\
        Avg. Intra-Cluster Distance & 20.53 & - \\
        Avg. Inter-Centroid Distance & 34.25 & - \\
        \bottomrule
    \end{tabular}
\end{table}

\begin{table}[h]
    \centering
    \small
    \caption{\textbf{Pairwise Euclidean distances between cluster centroids.} Cluster indices correspond to the colors in \cref{fig:latent_structure}: 1 (Blue), 2 (Orange), 3 (Green), 4 (Red), 5 (Purple). Note that the \textbf{Green-Orange} pair (bolded), despite being geometrically adjacent in the maze, exhibits the largest latent separation due to the wall barrier.}
    \label{tab:dist_matrix}
    \begin{tabular}{lccccc}
        \toprule
        Cluster & 1 (Blue) & 2 (Orange) & 3 (Green) & 4 (Red) & 5 (Purple) \\
        \midrule
        1 (Blue)   & 0     & 37.07 & 21.59 & 24.05 & 31.75 \\
        2 (Orange) & 37.07 & 0     & \textbf{53.21} & 23.67 & 28.45 \\
        3 (Green)  & 21.59 & \textbf{53.21} & 0     & 34.38 & 50.52 \\
        4 (Red)    & 24.05 & 23.67 & 34.38 & 0     & 37.89 \\
        5 (Purple) & 31.75 & 28.45 & 50.52 & 37.89 & 0     \\
        \bottomrule
    \end{tabular}
\end{table}

%% file: iclr2026_conference.bib
@article{kingma2016improved,
  title={Improved variational inference with inverse autoregressive flow},
  author={Kingma, Durk P and Salimans, Tim and Jozefowicz, Rafal and Chen, Xi and Sutskever, Ilya and Welling, Max},
  journal={Advances in neural information processing systems},
  volume={29},
  year={2016}
}

@article{luo2025learning,
  title={Learning to Trust Bellman Updates: Selective State-Adaptive Regularization for Offline RL},
  author={Luo, Qin-Wen and Xie, Ming-Kun and Wang, Ye-Wen and Huang, Sheng-Jun},
  journal={arXiv preprint arXiv:2505.19923},
  year={2025}
}

@article{tolman1948cognitive,
  title={Cognitive maps in rats and men.},
  author={Tolman, Edward C},
  journal={Psychological review},
  volume={55},
  number={4},
  pages={189},
  year={1948},
  publisher={American Psychological Association}
}

@article{whittington2022build,
  title={How to build a cognitive map},
  author={Whittington, James CR and McCaffary, David and Bakermans, Jacob JW and Behrens, Timothy EJ},
  journal={Nature neuroscience},
  volume={25},
  number={10},
  pages={1257--1272},
  year={2022},
  publisher={Nature Publishing Group US New York}
}

@article{gurnee2023language,
  title={Language models represent space and time},
  author={Gurnee, Wes and Tegmark, Max},
  journal={arXiv preprint arXiv:2310.02207},
  year={2023}
}

@article{vafa2024evaluating,
  title={Evaluating the world model implicit in a generative model},
  author={Vafa, Keyon and Chen, Justin Y and Rambachan, Ashesh and Kleinberg, Jon and Mullainathan, Sendhil},
  journal={Advances in Neural Information Processing Systems},
  volume={37},
  pages={26941--26975},
  year={2024}
}

@article{zhao2025place,
  title={Place Cells as Proximity-Preserving Embeddings: From Multi-Scale Random Walk to Straight-Forward Path Planning},
  author={Zhao, Minglu and Xu, Dehong and Kong, Deqian and Zhang, Wen-Hao and Wu, Ying Nian},
  journal={Advances in Neural Information Processing Systems},
  year={2025}
}

@article{o1976place,
  title={Place units in the hippocampus of the freely moving rat},
  author={O'Keefe, John},
  journal={Experimental neurology},
  volume={51},
  number={1},
  pages={78--109},
  year={1976},
  publisher={Elsevier}
}

@article{schulman2017proximal,
  title={Proximal policy optimization algorithms},
  author={Schulman, John and Wolski, Filip and Dhariwal, Prafulla and Radford, Alec and Klimov, Oleg},
  journal={arXiv preprint arXiv:1707.06347},
  year={2017}
}

@article{fujimoto2021minimalist,
  title={A minimalist approach to offline reinforcement learning},
  author={Fujimoto, Scott and Gu, Shixiang Shane},
  journal={Advances in neural information processing systems},
  volume={34},
  pages={20132--20145},
  year={2021}
}

@article{kumar2020conservative,
  title={Conservative q-learning for offline reinforcement learning},
  author={Kumar, Aviral and Zhou, Aurick and Tucker, George and Levine, Sergey},
  journal={Advances in neural information processing systems},
  volume={33},
  pages={1179--1191},
  year={2020}
}

@article{kostrikov2021offline,
  title={Offline reinforcement learning with implicit q-learning},
  author={Kostrikov, Ilya and Nair, Ashvin and Levine, Sergey},
  journal={arXiv preprint arXiv:2110.06169},
  year={2021}
}

@inproceedings{akiba2019optuna,
  title={Optuna: A next-generation hyperparameter optimization framework},
  author={Akiba, Takuya and Sano, Shotaro and Yanase, Toshihiko and Ohta, Takeru and Koyama, Masanori},
  booktitle={Proceedings of the 25th ACM SIGKDD international conference on knowledge discovery \& data mining},
  pages={2623--2631},
  year={2019}
}

@inproceedings{li2023overcoming,
  title={Overcoming posterior collapse in variational autoencoders via EM-type training},
  author={Li, Ying and Cheng, Lei and Yin, Feng and Zhang, Michael Minyi and Theodoridis, Sergios},
  booktitle={ICASSP 2023-2023 IEEE International Conference on Acoustics, Speech and Signal Processing (ICASSP)},
  pages={1--5},
  year={2023},
  organization={IEEE}
}

@article{kong2024latent,
  title   = {Latent plan transformer for trajectory abstraction: Planning as latent space inference},
  author  = {Kong, Deqian and Xu, Dehong and Zhao, Minglu and Pang, Bo and Xie, Jianwen and Lizarraga, Andrew and Huang, Yuhao and Xie, Sirui and Wu, Ying Nian},
  journal = {Advances in Neural Information Processing Systems},
  volume  = {37},
  pages   = {123379--123401},
  year    = {2024}
}

@article{chen2021decision,
  title   = {Decision transformer: reinforcement learning via sequence modeling, 2 June 2021},
  author  = {Chen, Lili and Lu, Kevin and Rajeswaran, Aravind and Lee, Kimin and Grover, Aditya and Laskin, Michael and Abbeel, Pieter and Srinivas, Aravind and Mordatch, Igor},
  journal = {URL http://arxiv. org/abs/2106.01345},
  year    = {2021}
}

@inproceedings{
kong2025latent,
title={Latent Thought Models with Variational Bayes Inference-Time Computation},
author={Deqian Kong and Minglu Zhao and Dehong Xu and Bo Pang and Shu Wang and Edouardo Honig and Zhangzhang Si and Chuan Li and Jianwen Xie and Sirui Xie and Ying Nian Wu},
booktitle={Forty-second International Conference on Machine Learning},
year={2025}
}

@article{nair2020awac,
  title={Awac: Accelerating online reinforcement learning with offline datasets},
  author={Nair, Ashvin and Gupta, Abhishek and Dalal, Murtaza and Levine, Sergey},
  journal={arXiv preprint arXiv:2006.09359},
  year={2020}
}

@article{wu2019behavior,
  title={Behavior regularized offline reinforcement learning},
  author={Wu, Yifan and Tucker, George and Nachum, Ofir},
  journal={arXiv preprint arXiv:1911.11361},
  year={2019}
}

@inproceedings{lee2022offline,
  title={Offline-to-online reinforcement learning via balanced replay and pessimistic q-ensemble},
  author={Lee, Seunghyun and Seo, Younggyo and Lee, Kimin and Abbeel, Pieter and Shin, Jinwoo},
  booktitle={Conference on Robot Learning},
  pages={1702--1712},
  year={2022},
  organization={PMLR}
}

@article{lee2022multi,
  title={Multi-game decision transformers},
  author={Lee, Kuang-Huei and Nachum, Ofir and Yang, Mengjiao Sherry and Lee, Lisa and Freeman, Daniel and Guadarrama, Sergio and Fischer, Ian and Xu, Winnie and Jang, Eric and Michalewski, Henryk and others},
  journal={Advances in Neural Information Processing Systems},
  volume={35},
  pages={27921--27936},
  year={2022}
}

@inproceedings{yamagata2023q,
  title={Q-learning decision transformer: Leveraging dynamic programming for conditional sequence modelling in offline rl},
  author={Yamagata, Taku and Khalil, Ahmed and Santos-Rodriguez, Raul},
  booktitle={International Conference on Machine Learning},
  pages={38989--39007},
  year={2023},
  organization={PMLR}
}

@article{kumar2019stabilizing,
  title={Stabilizing off-policy q-learning via bootstrapping error reduction},
  author={Kumar, Aviral and Fu, Justin and Soh, Matthew and Tucker, George and Levine, Sergey},
  journal={Advances in neural information processing systems},
  volume={32},
  year={2019}
}

@inproceedings{bellemare2017distributional,
  title={A distributional perspective on reinforcement learning},
  author={Bellemare, Marc G and Dabney, Will and Munos, R{\'e}mi},
  booktitle={International conference on machine learning},
  pages={449--458},
  year={2017},
  organization={PMLR}
}

@inproceedings{huang2023reparameterized,
  title={Reparameterized policy learning for multimodal trajectory optimization},
  author={Huang, Zhiao and Liang, Litian and Ling, Zhan and Li, Xuanlin and Gan, Chuang and Su, Hao},
  booktitle={International Conference on Machine Learning},
  pages={13957--13975},
  year={2023},
  organization={PMLR}
}

@article{jordan1999introduction,
  title={An introduction to variational methods for graphical models},
  author={Jordan, Michael I and Ghahramani, Zoubin and Jaakkola, Tommi S and Saul, Lawrence K},
  journal={Machine learning},
  volume={37},
  pages={183--233},
  year={1999},
  publisher={Springer}
}

@article{blei2017variational,
  title={Variational inference: A review for statisticians},
  author={Blei, David M and Kucukelbir, Alp and McAuliffe, Jon D},
  journal={Journal of the American statistical Association},
  volume={112},
  number={518},
  pages={859--877},
  year={2017},
  publisher={Taylor \& Francis}
}

@article{murphy2018machine,
  title={Machine learning: A probabilistic perspective (adaptive computation and machine learning series)},
  author={Murphy, Kevin P},
  journal={The MIT Press: London, UK},
  year={2018}
}

@book{sutton1998reinforcement,
  title={Reinforcement learning: An introduction},
  author={Sutton, Richard S and Barto, Andrew G and others},
  volume={1},
  number={1},
  year={1998},
  publisher={MIT press Cambridge}
}

@inproceedings{fujimoto2019off,
  title={Off-policy deep reinforcement learning without exploration},
  author={Fujimoto, Scott and Meger, David and Precup, Doina},
  booktitle={International conference on machine learning},
  pages={2052--2062},
  year={2019},
  organization={PMLR}
}

@article{wu2022supported,
  title={Supported policy optimization for offline reinforcement learning},
  author={Wu, Jialong and Wu, Haixu and Qiu, Zihan and Wang, Jianmin and Long, Mingsheng},
  journal={Advances in Neural Information Processing Systems},
  volume={35},
  pages={31278--31291},
  year={2022}
}

@article{lyu2022mildly,
  title={Mildly conservative q-learning for offline reinforcement learning},
  author={Lyu, Jiafei and Ma, Xiaoteng and Li, Xiu and Lu, Zongqing},
  journal={Advances in Neural Information Processing Systems},
  volume={35},
  pages={1711--1724},
  year={2022}
}

@inproceedings{zheng2022online,
  title={Online decision transformer},
  author={Zheng, Qinqing and Zhang, Amy and Grover, Aditya},
  booktitle={international conference on machine learning},
  pages={27042--27059},
  year={2022},
  organization={PMLR}
}

@inproceedings{ronneberger2015u,
  title={U-net: Convolutional networks for biomedical image segmentation},
  author={Ronneberger, Olaf and Fischer, Philipp and Brox, Thomas},
  booktitle={Medical image computing and computer-assisted intervention--MICCAI 2015: 18th international conference, Munich, Germany, October 5-9, 2015, proceedings, part III 18},
  pages={234--241},
  year={2015},
  organization={Springer}
}

@article{hoffman2013stochastic,
  title={Stochastic variational inference},
  author={Hoffman, Matthew D and Blei, David M and Wang, Chong and Paisley, John},
  journal={the Journal of machine Learning research},
  volume={14},
  number={1},
  pages={1303--1347},
  year={2013},
  publisher={JMLR. org}
}

@article{kingma2013auto,
  title={Auto-encoding variational bayes},
  author={Kingma, Diederik P and Welling, Max},
  journal={International Conference on Learning Representations},
  year={2014}
}

@article{kidambi2020morel,
  title={Morel: Model-based offline reinforcement learning},
  author={Kidambi, Rahul and Rajeswaran, Aravind and Netrapalli, Praneeth and Joachims, Thorsten},
  journal={Advances in neural information processing systems},
  volume={33},
  pages={21810--21823},
  year={2020}
}

@article{yu2020mopo,
  title={Mopo: Model-based offline policy optimization},
  author={Yu, Tianhe and Thomas, Garrett and Yu, Lantao and Ermon, Stefano and Zou, James Y and Levine, Sergey and Finn, Chelsea and Ma, Tengyu},
  journal={Advances in Neural Information Processing Systems},
  volume={33},
  pages={14129--14142},
  year={2020}
}

@article{peng2019advantage,
  title={Advantage-weighted regression: Simple and scalable off-policy reinforcement learning},
  author={Peng, Xue Bin and Kumar, Aviral and Zhang, Grace and Levine, Sergey},
  journal={arXiv preprint arXiv:1910.00177},
  year={2019}
}

@article{emmons2021rvs,
  title={Rvs: What is essential for offline rl via supervised learning?},
  author={Emmons, Scott and Eysenbach, Benjamin and Kostrikov, Ilya and Levine, Sergey},
  journal={arXiv preprint arXiv:2112.10751},
  year={2021}
}

@inproceedings{rosete2023latent,
  title={Latent plans for task-agnostic offline reinforcement learning},
  author={Rosete-Beas, Erick and Mees, Oier and Kalweit, Gabriel and Boedecker, Joschka and Burgard, Wolfram},
  booktitle={Conference on Robot Learning},
  pages={1838--1849},
  year={2023},
  organization={PMLR}
}

@article{painter2023monte,
  title={Monte carlo tree search with boltzmann exploration},
  author={Painter, Michael and Baioumy, Mohamed and Hawes, Nick and Lacerda, Bruno},
  journal={Advances in Neural Information Processing Systems},
  volume={36},
  pages={78181--78192},
  year={2023}
}

@inproceedings{zhou2020neural,
  title={Neural contextual bandits with ucb-based exploration},
  author={Zhou, Dongruo and Li, Lihong and Gu, Quanquan},
  booktitle={International Conference on Machine Learning},
  pages={11492--11502},
  year={2020},
  organization={PMLR}
}

@inproceedings{wang2022thompson,
  title={Thompson sampling for (combinatorial) pure exploration},
  author={Wang, Siwei and Zhu, Jun},
  booktitle={International Conference on Machine Learning},
  pages={23470--23483},
  year={2022},
  organization={PMLR}
}

@article{ecoffet2019go,
  title={Go-explore: a new approach for hard-exploration problems},
  author={Ecoffet, Adrien and Huizinga, Joost and Lehman, Joel and Stanley, Kenneth O and Clune, Jeff},
  journal={arXiv preprint arXiv:1901.10995},
  year={2019}
}

@article{fortunato2017noisy,
  title={Noisy networks for exploration},
  author={Fortunato, Meire and Azar, Mohammad Gheshlaghi and Piot, Bilal and Menick, Jacob and Osband, Ian and Graves, Alex and Mnih, Vlad and Munos, Remi and Hassabis, Demis and Pietquin, Olivier and others},
  journal={arXiv preprint arXiv:1706.10295},
  year={2017}
}

@article{fu2017ex2,
  title={Ex2: Exploration with exemplar models for deep reinforcement learning},
  author={Fu, Justin and Co-Reyes, John and Levine, Sergey},
  journal={Advances in neural information processing systems},
  volume={30},
  year={2017}
}

@article{bellemare2016unifying,
  title={Unifying count-based exploration and intrinsic motivation},
  author={Bellemare, Marc and Srinivasan, Sriram and Ostrovski, Georg and Schaul, Tom and Saxton, David and Munos, Remi},
  journal={Advances in neural information processing systems},
  volume={29},
  year={2016}
}

@article{tang2017exploration,
  title={\# exploration: A study of count-based exploration for deep reinforcement learning},
  author={Tang, Haoran and Houthooft, Rein and Foote, Davis and Stooke, Adam and Xi Chen, OpenAI and Duan, Yan and Schulman, John and DeTurck, Filip and Abbeel, Pieter},
  journal={Advances in neural information processing systems},
  volume={30},
  year={2017}
}

@article{burda2018exploration,
  title={Exploration by random network distillation},
  author={Burda, Yuri and Edwards, Harrison and Storkey, Amos and Klimov, Oleg},
  journal={arXiv preprint arXiv:1810.12894},
  year={2018}
}

@article{badia2020never,
  title={Never give up: Learning directed exploration strategies},
  author={Badia, Adri{\`a} Puigdom{\`e}nech and Sprechmann, Pablo and Vitvitskyi, Alex and Guo, Daniel and Piot, Bilal and Kapturowski, Steven and Tieleman, Olivier and Arjovsky, Mart{\'\i}n and Pritzel, Alexander and Bolt, Andew and others},
  journal={arXiv preprint arXiv:2002.06038},
  year={2020}
}

@inproceedings{badia2020agent57,
  title={Agent57: Outperforming the atari human benchmark},
  author={Badia, Adri{\`a} Puigdom{\`e}nech and Piot, Bilal and Kapturowski, Steven and Sprechmann, Pablo and Vitvitskyi, Alex and Guo, Zhaohan Daniel and Blundell, Charles},
  booktitle={International conference on machine learning},
  pages={507--517},
  year={2020},
  organization={PMLR}
}

@inproceedings{haarnoja2018soft,
  title={Soft actor-critic: Off-policy maximum entropy deep reinforcement learning with a stochastic actor},
  author={Haarnoja, Tuomas and Zhou, Aurick and Abbeel, Pieter and Levine, Sergey},
  booktitle={International conference on machine learning},
  pages={1861--1870},
  year={2018},
  organization={Pmlr}
}

@article{houthooft2016vime,
  title={Vime: Variational information maximizing exploration},
  author={Houthooft, Rein and Chen, Xi and Duan, Yan and Schulman, John and De Turck, Filip and Abbeel, Pieter},
  journal={Advances in neural information processing systems},
  volume={29},
  year={2016}
}

@article{eysenbach2018diversity,
  title={Diversity is all you need: Learning skills without a reward function},
  author={Eysenbach, Benjamin and Gupta, Abhishek and Ibarz, Julian and Levine, Sergey},
  journal={arXiv preprint arXiv:1802.06070},
  year={2018}
}

@article{beeson2022improving,
  title={Improving td3-bc: Relaxed policy constraint for offline learning and stable online fine-tuning},
  author={Beeson, Alex and Montana, Giovanni},
  journal={arXiv preprint arXiv:2211.11802},
  year={2022}
}

@article{nakamoto2023cal,
  title={Cal-ql: Calibrated offline rl pre-training for efficient online fine-tuning},
  author={Nakamoto, Mitsuhiko and Zhai, Simon and Singh, Anikait and Sobol Mark, Max and Ma, Yi and Finn, Chelsea and Kumar, Aviral and Levine, Sergey},
  journal={Advances in Neural Information Processing Systems},
  volume={36},
  pages={62244--62269},
  year={2023}
}

@article{luo2023finetuning,
  title={Finetuning from offline reinforcement learning: Challenges, trade-offs and practical solutions},
  author={Luo, Yicheng and Kay, Jackie and Grefenstette, Edward and Deisenroth, Marc Peter},
  journal={arXiv preprint arXiv:2303.17396},
  year={2023}
}

@article{zhang2023policy,
  title={Policy expansion for bridging offline-to-online reinforcement learning},
  author={Zhang, Haichao and Xu, We and Yu, Haonan},
  journal={arXiv preprint arXiv:2302.00935},
  year={2023}
}

@article{li2023proto,
  title={Proto: Iterative policy regularized offline-to-online reinforcement learning},
  author={Li, Jianxiong and Hu, Xiao and Xu, Haoran and Liu, Jingjing and Zhan, Xianyuan and Zhang, Ya-Qin},
  journal={arXiv preprint arXiv:2305.15669},
  year={2023}
}

@article{liu2024energy,
  title={Energy-guided diffusion sampling for offline-to-online reinforcement learning},
  author={Liu, Xu-Hui and Liu, Tian-Shuo and Jiang, Shengyi and Chen, Ruifeng and Zhang, Zhilong and Chen, Xinwei and Yu, Yang},
  journal={arXiv preprint arXiv:2407.12448},
  year={2024}
}

@article{janner2021offline,
  title={Offline reinforcement learning as one big sequence modeling problem},
  author={Janner, Michael and Li, Qiyang and Levine, Sergey},
  journal={Advances in neural information processing systems},
  volume={34},
  pages={1273--1286},
  year={2021}
}

@article{chi2023diffusion,
  title={Diffusion policy: Visuomotor policy learning via action diffusion},
  author={Chi, Cheng and Xu, Zhenjia and Feng, Siyuan and Cousineau, Eric and Du, Yilun and Burchfiel, Benjamin and Tedrake, Russ and Song, Shuran},
  journal={The International Journal of Robotics Research},
  pages={02783649241273668},
  year={2023},
  publisher={SAGE Publications Sage UK: London, England}
}

@inproceedings{janner2022planning,
  title={Planning with Diffusion for Flexible Behavior Synthesis},
  author={Janner, Michael and Du, Yilun and Tenenbaum, Joshua and Levine, Sergey},
  booktitle={International Conference on Machine Learning},
  pages={9902--9915},
  year={2022},
  organization={PMLR}
}

@article{zheng2023guided,
  title={Guided flows for generative modeling and decision making},
  author={Zheng, Qinqing and Le, Matt and Shaul, Neta and Lipman, Yaron and Grover, Aditya and Chen, Ricky TQ},
  journal={arXiv preprint arXiv:2311.13443},
  year={2023}
}

@article{lipman2022flow,
  title={Flow matching for generative modeling},
  author={Lipman, Yaron and Chen, Ricky TQ and Ben-Hamu, Heli and Nickel, Maximilian and Le, Matt},
  journal={arXiv preprint arXiv:2210.02747},
  year={2022}
}

@article{yang2022dichotomy,
  title={Dichotomy of control: Separating what you can control from what you cannot},
  author={Yang, Mengjiao and Schuurmans, Dale and Abbeel, Pieter and Nachum, Ofir},
  journal={arXiv preprint arXiv:2210.13435},
  year={2022}
}

@article{paster2022you,
  title={You can’t count on luck: Why decision transformers and rvs fail in stochastic environments},
  author={Paster, Keiran and McIlraith, Sheila and Ba, Jimmy},
  journal={Advances in neural information processing systems},
  volume={35},
  pages={38966--38979},
  year={2022}
}

@article{fu2020d4rl,
  title={D4rl: Datasets for deep data-driven reinforcement learning},
  author={Fu, Justin and Kumar, Aviral and Nachum, Ofir and Tucker, George and Levine, Sergey},
  journal={arXiv preprint arXiv:2004.07219},
  year={2020}
}

@article{levine2020offline,
  title={Offline reinforcement learning: Tutorial, review, and perspectives on open problems},
  author={Levine, Sergey and Kumar, Aviral and Tucker, George and Fu, Justin},
  journal={arXiv preprint arXiv:2005.01643},
  year={2020}
}

@article{konda1999actor,
  title={Actor-critic algorithms},
  author={Konda, Vijay and Tsitsiklis, John},
  journal={Advances in neural information processing systems},
  volume={12},
  year={1999}
}

@inproceedings{barth2018distributed,
  title={Distributed Distributional Deterministic Policy Gradients},
  author={Barth-Maron, Gabriel and Hoffman, Matthew W and Budden, David and Dabney, Will and Horgan, Dan and Dhruva, TB and Muldal, Alistair and Heess, Nicolas and Lillicrap, Timothy},
  booktitle={International Conference on Learning Representations},
  year={2018}
}

@inproceedings{gruslys2018reactor,
  title={The Reactor: A fast and sample-efficient Actor-Critic agent for Reinforcement Learning},
  author={Gruslys, Audrunas and Dabney, Will and Azar, Mohammad Gheshlaghi and Piot, Bilal and Bellemare, Marc and Munos, Remi},
  booktitle={International Conference on Learning Representations},
  year={2018}
}

@inproceedings{ajay2023conditional,
  title={Is Conditional Generative Modeling all you need for Decision Making?},
  author={Ajay, Anurag and Du, Yilun and Gupta, Abhi and Tenenbaum, Joshua B and Jaakkola, Tommi S and Agrawal, Pulkit},
  booktitle={The Eleventh International Conference on Learning Representations},
  year={2023},
}

@article{mnih2015human,
  title={Human-level control through deep reinforcement learning},
  author={Mnih, Volodymyr and Kavukcuoglu, Koray and Silver, David and Rusu, Andrei A and Veness, Joel and Bellemare, Marc G and Graves, Alex and Riedmiller, Martin and Fidjeland, Andreas K and Ostrovski, Georg and others},
  journal={nature},
  volume={518},
  number={7540},
  pages={529--533},
  year={2015},
  publisher={Nature Publishing Group}
}

@article{schaul2015prioritized,
  title={Prioritized experience replay},
  author={Schaul, Tom and Quan, John and Antonoglou, Ioannis and Silver, David},
  journal={arXiv preprint arXiv:1511.05952},
  year={2015}
}

@book{ziebart2010modeling,
  title={Modeling purposeful adaptive behavior with the principle of maximum causal entropy},
  author={Ziebart, Brian D},
  year={2010},
  publisher={Carnegie Mellon University}
}

@inproceedings{fox2016taming,
  title={Taming the noise in reinforcement learning via soft updates},
  author={Fox, Roy and Pakman, Ari and Tishby, Naftali},
  booktitle={32nd Conference on Uncertainty in Artificial Intelligence 2016, UAI 2016},
  pages={202--211},
  year={2016},
  organization={Association For Uncertainty in Artificial Intelligence (AUAI)}
}

@inproceedings{silver2014deterministic,
  title={Deterministic policy gradient algorithms},
  author={Silver, David and Lever, Guy and Heess, Nicolas and Degris, Thomas and Wierstra, Daan and Riedmiller, Martin},
  booktitle={International conference on machine learning},
  pages={387--395},
  year={2014},
  organization={Pmlr}
}

@book{mccallum1996reinforcement,
  title={Reinforcement learning with selective perception and hidden state},
  author={McCallum, Andrew Kachites},
  year={1996},
  publisher={University of Rochester}
}

@article{srivastava2019training,
  title={Training agents using upside-down reinforcement learning},
  author={Srivastava, Rupesh Kumar and Shyam, Pranav and Mutz, Filipe and Ja{\'s}kowski, Wojciech and Schmidhuber, J{\"u}rgen},
  journal={arXiv preprint arXiv:1912.02877},
  year={2019}
}

@article{levine2018reinforcement,
  title={Reinforcement learning and control as probabilistic inference: Tutorial and review},
  author={Levine, Sergey},
  journal={arXiv preprint arXiv:1805.00909},
  year={2018}
}

@article{botvinick2012planning,
  title={Planning as inference},
  author={Botvinick, Matthew and Toussaint, Marc},
  journal={Trends in cognitive sciences},
  volume={16},
  number={10},
  pages={485--488},
  year={2012},
  publisher={Elsevier}
}

@inproceedings{abdolmaleki2018maximum,
  title={Maximum a Posteriori Policy Optimisation},
  author={Abdolmaleki, Abbas and Springenberg, Jost Tobias and Tassa, Yuval and Munos, Remi and Heess, Nicolas and Riedmiller, Martin},
  booktitle={International Conference on Learning Representations},
  year={2018}
}

@article{qin2023learning,
  title={Learning non-Markovian decision-making from state-only sequences},
  author={Qin, Aoyang and Gao, Feng and Li, Qing and Zhu, Song-Chun and Xie, Sirui},
  journal={Advances in Neural Information Processing Systems},
  volume={36},
  pages={6596--6618},
  year={2023}
}

@article{rubinstein1999cross,
  title={The cross-entropy method for combinatorial and continuous optimization},
  author={Rubinstein, Reuven},
  journal={Methodology and computing in applied probability},
  volume={1},
  number={2},
  pages={127--190},
  year={1999},
  publisher={Springer}
}

@article{abel2021expressivity,
  title={On the expressivity of markov reward},
  author={Abel, David and Dabney, Will and Harutyunyan, Anna and Ho, Mark K and Littman, Michael and Precup, Doina and Singh, Satinder},
  journal={Advances in Neural Information Processing Systems},
  volume={34},
  pages={7799--7812},
  year={2021}
}

@inproceedings{bowling2023settling,
  title={Settling the reward hypothesis},
  author={Bowling, Michael and Martin, John D and Abel, David and Dabney, Will},
  booktitle={International Conference on Machine Learning},
  pages={3003--3020},
  year={2023},
  organization={PMLR}
}
